\documentclass[preprint,10pt]{elsarticle}
\usepackage[margin=1in]{geometry}
\usepackage{amsmath}
\usepackage{comment}
\usepackage{amssymb}
\usepackage{amsfonts}
\usepackage{graphicx}
\usepackage{textcomp}
\usepackage{xcolor}
\usepackage{booktabs}
\usepackage{tabularx}
\usepackage{longtable}
\usepackage{multirow}
\usepackage{array}
\usepackage{rotating}
\usepackage{subcaption}
\usepackage{adjustbox}
\usepackage{threeparttable}
\usepackage[referable]{threeparttablex}
\usepackage[hidelinks]{hyperref}
\hypersetup{
  pdftitle={Toward Deployable Bangla Sign Language Recognition with Expert-Validated Data and a Lightweight Attention-Based Model},
  pdfauthor={Saad Ahmed and Md Khalid Syfullah}}
\usepackage{fontspec}  
\newfontfamily\bnfont{Kalpurush.ttf}[Script=Bengali]
\newcommand{\bn}[1]{{\bnfont #1}}
\biboptions{sort&compress}
\renewcommand{\arraystretch}{1.15}
\journal{Computer Vision and Image Understanding}
\begin{document}
\begin{frontmatter}
\title{Toward Deployable Bangla Sign Language Recognition with Expert-Validated Data and a Lightweight Attention-Based Model}
\author[1]{Saad Ahmed\corref{cor1}}
\author[1]{Md Khalid Syfullah}
\affiliation[1]{organization={Department of Computer Science and Engineering, Bangladesh Army University of Science and Technology}, addressline={Saidpur Cantonment}, city={Saidpur}, country={Bangladesh}}
\begin{abstract}

Deaf and hard-of-hearing people in Bangladesh communicate mainly through Bangla Sign Language (BdSL). Automatic BdSL recognition on personal devices could widen access to education and services. Existing systems use controlled-setting datasets without expert verification and heavyweight pretrained backbones unsuited to on-device use. We introduce RSBdSL38, 10,874 expert-validated images spanning all 38 BdSL hand signs, representing the 51 letters of the Bangla alphabet, recorded from real signers at three special-needs schools across Bangladesh. We propose a lightweight attention-based convolutional network of 298,470 parameters, built from grouped bottleneck residual blocks, channel and spatial attention, a multi-scale depthwise hand-feature block, dual pooling, and Swish activations. Trained from scratch, it attains 96.37\% accuracy (95.72\% $\pm$ 0.54\% over five seeds), within 1.08 percentage points of the best of nine ImageNet-pretrained efficient architectures under an identical protocol, using 8.5 to 68$\times$ fewer parameters and 1.3 to 21.7$\times$ fewer MACs. Retrained, it reaches 92.95 to 98.33\% on six public BdSL benchmarks, 97.04\% on a merged corpus, and 76.25\% zero-shot on BdSL-38. Removing any architectural stage costs 7.61 to 89.30 points, against at most 3.17 for the training recipe. Grad-CAM with deletion--insertion and weight-randomization checks confirms that predictions follow the signing hand. A signer-independent split holding out 6 of 36 signers yields 85.18\%. Quantized to 0.48~MB, it runs at 3.98~ms per image within a 15.5~MB footprint on a commodity smartphone. Together, RSBdSL38 and our from-scratch model turn benchmark accuracy into deployable accessibility at a fraction of pretrained-backbone cost; dataset, code, and models are released.
\end{abstract}


\end{frontmatter}
\section{Introduction}
\label{sec:introduction}
Spoken language is inaccessible to the approximately 430 million people worldwide, over 5\% of the global population, who live with disabling hearing loss, a figure the World Health Organization projects will exceed 700 million by 2050 \cite{who2023deafness}. In Bangladesh the prevalence of hearing impairment is nearly 9.6\%, almost double the global average, and an estimated 2.5 to 3 million people rely on Bangla Sign Language (BdSL) daily \cite{tarafder2015disabling,hadiuzzaman2024baust}. Their barriers to education, healthcare, employment, and social participation follow not from the disability itself but from the absence of communication bridges. Certified BdSL interpreters are few and concentrated in urban centres, so for most of the deaf and hard-of-hearing (DHH) population an interpreter is not available when communication is needed. An automatic recognition system running on the hardware people already own is therefore not a convenience but a substitute for a service that does not otherwise exist.\par
Sign language is a complete, rule-governed visual-gestural language with its own phonology, morphology, grammar, and lexicon, and is not universally shared across nations or regions \cite{valli2000linguistics,sutton1999linguistics}. BdSL, formally recognized in 2000 by the Centre for Disability in Development, is the primary means of communication for the DHH community in Bangladesh \cite{kabir2025combining}; it comprises 38 hand signs representing the 51 letters of the Bangla alphabet, in both one-handed and two-handed forms \cite{nihal2021bangla}. Alphabet-level, or fingerspelling, recognition is the entry point to the problem: it underpins name and loanword spelling, classroom instruction, and interactive learning tools, and supplies the visual front end on which later word- and sentence-level systems are built \cite{rastgoo2021sign}.\par
Visually the task is harder than its 38-way label space suggests. Several BdSL signs differ only in the flexion of a single finger or a small change of palm orientation, so the between-class distance in image space can be smaller than the within-class distance induced by a change of signer, viewpoint, or lighting; skin tone, hand size, and articulation habits all vary strongly across signers. The discriminative evidence occupies a small, deformable, self-occluding region, while the remaining pixels, clothing, furniture, classroom walls, are class-irrelevant and often higher in contrast than the hand itself. A recognizer that works outside the laboratory must therefore learn where to look as well as what to look for, from the modest labelled data that low-resource sign languages can realistically provide.\par
Deep learning has advanced automatic sign language recognition (SLR) considerably. Convolutional Neural Networks (CNNs) are the dominant paradigm for image-based SLR, reaching high accuracy on American, Chinese, and German Sign Language \cite{al2021deep,renjith2024sign}; transfer learning from large pretrained models is widely used to compensate for the limited data available in low-resource domains \cite{haque2023recognition,oquab2023dinov2,tasnim2026vision}, alongside ensembles, attention mechanisms, and hybrid CNN-LSTM designs \cite{kabir2025combining,hadiuzzaman2024baust,podder2022bangla}. In parallel, the wider vision community has produced a mature family of efficient backbones, MobileNetV4, MobileViT, EfficientNetV2, EfficientFormerV2, and GhostNetV2 among them \cite{qin2024mobilenetv4,mehta2022mobilevit,mehta2022separable,tan2021efficientnetv2,li2023rethinking,tang2022ghostnetv2}, attention modules that improve feature selectivity at small parameter cost \cite{hu2018squeeze,woo2018cbam}, and attribution tools that expose what a trained model actually uses \cite{selvaraju2017grad,chattopadhay2018grad,petsiuk2018rise,adebayo2018sanity}. These three lines, efficient design, attention, and verifiable explanation, are individually well developed but are rarely brought together in the BdSL literature.\par
Deployability is the property that decides whether any of this reaches a user. An assistive BdSL recognizer will most plausibly run offline on a mid-range Android phone, a classroom tablet, or a low-cost embedded board, sharing memory and battery with the application around it. A 20M-parameter backbone executing hundreds of millions of multiply--accumulate operations per frame is a poor fit under that constraint, however strong its benchmark accuracy: accuracy per parameter, per multiply--accumulate operation, and per megabyte matters as much as accuracy itself, and should be measured on the target hardware rather than inferred from parameter counts alone.\par
Despite this broader progress, BdSL recognition remains comparatively underexplored, for three reasons. First, expert-verified BdSL data is scarce: the existing public datasets, BDSL49 \cite{hasib2023bdsl}, BdSL36 \cite{hoque2020bdsl36}, KU-BdSL \cite{jim2023ku}, BAUST Lipi \cite{hadiuzzaman2024baust}, BdSL47 \cite{rayeed2023bdsl47}, and Shongket \cite{hasan2021shongket}, share few contributors (typically 10 to 42), predominantly non-signing volunteers, no complete expert linguistic verification (BdSL36 \cite{hoque2020bdsl36} being a partial exception, collected at a deaf school and filtered by signer experts), and controlled capture that does not reflect real-world diversity, all of which constrain generalizability. Second, prior BdSL models are either heavyweight pretrained architectures not optimized for efficient deployment, or custom models evaluated only on their own data without cross-dataset or ablation validation \cite{islam2022sign,11013937}. Third, the evaluation protocols are permissive: results are almost universally reported on class-stratified splits in which images of the same signer appear in both partitions, so the reported figure silently includes what the model has learned about individual people, and neither signer-independent nor zero-shot cross-corpus performance is characterized. It therefore often remains unclear whether reported accuracies reflect architectural merit, dataset idiosyncrasies, or evaluation leakage.\par
To address these gaps, we propose \emph{RSBdSL38}, an expert-validated BdSL dataset collected from real signers, together with a lightweight attention-based recognition model designed for resource-constrained deployment, and an evaluation protocol that measures the properties deployment actually depends on. Our contributions are four-fold:
\begin{itemize}
    \item We introduce RSBdSL38, a BdSL dataset of 10,874 images covering all 38 BdSL hand signs (51 Bangla letters), collected from real signers at three special-needs schools in three regions of Bangladesh together with additional volunteers, and validated in full by a sign language expert to ensure linguistic authenticity.
    \item We design a lightweight attention-based CNN of only 298{,}470 parameters (1.14~MB, 132.7~M multiply--accumulate operations) integrating grouped bottleneck residual blocks, channel and spatial attention, a multi-scale depthwise \emph{hand-feature block}, dual-pooling aggregation, and Swish activations, which trained from scratch is competitive with nine modern ImageNet-pretrained efficient architectures at 8.5 to 68$\times$ fewer parameters and generalizes to six public BdSL benchmarks and a merged four-dataset corpus without architectural change.
    \item We validate the design through a stage-wise depth ablation covering all 14 stage-removal combinations and parameter-matched component controls that isolate the training recipe and the activation function from model capacity, and verify that the learned attention localizes the signing hand using Grad-CAM and Grad-CAM++ maps with deletion and insertion faithfulness measures and a weight-randomization sanity check.
    \item We report the deployment-facing evaluation the BdSL literature omits: a signer-independent protocol holding out 6 of the 36 signers entirely, zero-shot cross-corpus transfer without fine-tuning, and end-to-end efficiency measurements including multiply--accumulate cost, quantized model size, and on-device latency and memory on a commodity Android smartphone.
\end{itemize}
We investigate the following Research Questions (RQs):
\begin{itemize}
    \item \textbf{RQ1:} Can a lightweight model trained from scratch match modern ImageNet-pretrained efficient architectures on BdSL recognition at a fraction of the parameter budget?
    \item \textbf{RQ2:} How well does the proposed architecture generalize across public BdSL benchmarks and a merged multi-dataset corpus without any architectural modification, and how much of that performance survives when the trained model is transferred to a new corpus with no adaptation?
    \item \textbf{RQ3:} Which architectural components and training-recipe elements contribute most to recognition performance, as quantified by systematic ablation?
    \item \textbf{RQ4:} Does expert-validated data collected from real signers (RSBdSL38) constitute a more challenging and ecologically valid benchmark than existing volunteer-collected BdSL datasets, and how much accuracy is lost when the evaluation removes signer overlap?
    \item \textbf{RQ5:} Does the model base its predictions on the signing hand rather than on background context, as required for deployment in uncontrolled environments?
\end{itemize}
The paper is organized as follows. Section~\ref{sec:background} presents the background concepts and Section~\ref{sec:litreview} reviews related work and identifies the research gaps. Section~\ref{sec:dataset} introduces RSBdSL38, Section~\ref{sec:methodology} details the architecture and methodology, and Section~\ref{sec:results} reports the experiments and findings. Section~\ref{sec:conclusion} concludes; supporting material, including the detailed architecture diagram, the per-class classification report, and the complete ablation tables, is provided in the appendices.
\section{Background}
\label{sec:background}
\begin{figure}[!t]
    \centering
    \includegraphics[width=\columnwidth]{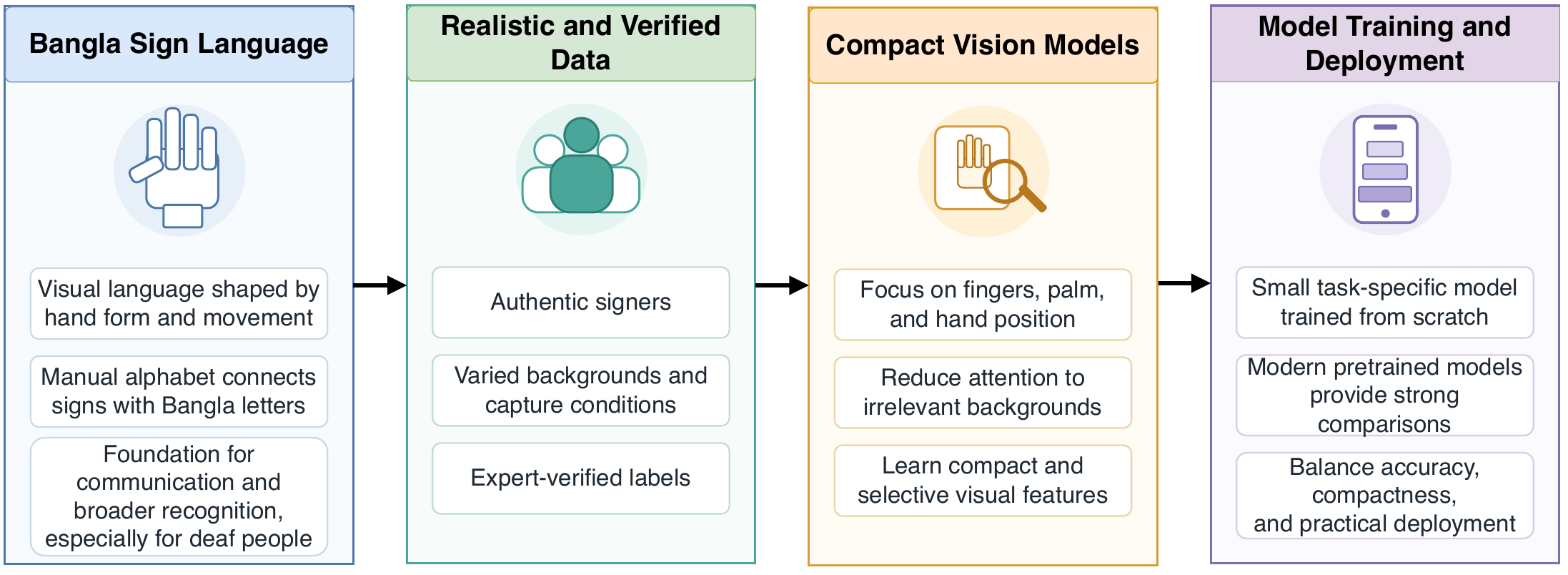}
    \caption{Conceptual overview of the background for Bangla Sign Language recognition: the structure of the BdSL manual alphabet, the data-collection requirements of a realistic benchmark, the role of channel and spatial attention in compact models, the building blocks of efficient convolutional design, and the choice between ImageNet-pretrained backbones and task-specific networks trained from scratch.}
    \label{fig:background}
\end{figure}

This section reviews the concepts the proposed recognizer builds on. Fig.~\ref{fig:background} summarizes how they relate to one another and to the contributions of this paper.\par
\subsection{Bangla Sign Language and Alphabet-Level Recognition}
BdSL is not a visual copy of spoken Bangla; it has its own signing conventions, hand shapes, and movement patterns. Its manual alphabet uses 38 canonical hand signs for the 51 letters of the written script, because multiple letters share an identical sign form \cite{nihal2021bangla,kabir2025combining}. Alphabet-level recognition is foundational, supporting fingerspelling, classroom communication, assistive translation, and later word- and sentence-level systems \cite{rastgoo2021sign}, but demanding: several classes are separated only by the position of a single finger, so the recognizer must resolve fine articulatory detail rather than coarse hand silhouette.\par
\subsection{Need for Realistic and Expert-Verified Data}
Recognizers are highly sensitive to how the data are collected: a model trained on posed gestures from a few volunteers can score well on a test split from the same setting and fail on real users in new environments. Expert verification matters for a specific reason: small changes in finger shape or hand orientation change a sign's meaning, and a volunteer who does not sign daily may produce a form close to, but not, the canonical one, which then enters the training set as label noise. A realistic benchmark therefore needs authentic signers, diverse capture conditions, expert validation, and a protocol that does not place one signer's images in both partitions.\par
\subsection{Attention in Lightweight Vision Models}
Attention mechanisms help a network focus on the most informative parts of an image. In hand-sign recognition the evidence is concentrated around the fingers, palm, and relative hand positions, while the background carries little class information: channel attention emphasizes useful feature types and spatial attention highlights important regions \cite{hu2018squeeze,woo2018cbam}. Both are cheap, a channel-attention module with reduction ratio $r$ adding only $2C^2/r$ weights to a $C$-channel feature map and a single-map spatial gate one convolution kernel, which makes them the cheapest way for a sub-megabyte network to decide where to spend capacity.\par
\subsection{Efficient Convolutional Design}
Efficient convolutional networks reduce computational cost by reusing simple building blocks: residual connections, bottleneck layers, grouped convolutions, and depthwise operations \cite{he2016deep,krizhevsky2012imagenet,xie2017aggregated,chollet2017xception}. A bottleneck halves the channel count before the spatial convolution and restores it afterwards; a grouped convolution with $g$ groups divides the cost of that layer by $g$; a depthwise convolution takes this to its limit by giving every channel its own kernel. These designs suit assistive technologies that must run on smartphones, classroom devices, or low-cost embedded hardware. Here the model is designed from the outset as a small BdSL-specific recognizer rather than a reduced version of a much larger one.\par
\subsection{Pretrained Backbones and Training from Scratch}
Transfer learning from ImageNet-pretrained models is common in sign language recognition because it often improves accuracy when target data are limited \cite{deng2009imagenet,sraboni2024real,kabir2025combining}, and modern efficient families such as MobileNetV4, MobileViT, EfficientNetV2, EfficientFormerV2, and GhostNetV2 offer strong baselines \cite{qin2024mobilenetv4,mehta2022mobilevit,mehta2022separable,tan2021efficientnetv2,li2023rethinking,tang2022ghostnetv2}. Pretrained backbones nonetheless demand more memory and infrastructure than a compact task-specific model, and their capacity is allocated to a 1000-class natural-image problem whose statistics differ substantially from close-range hand imagery, which is the trade-off between accuracy, compactness, and deployment practicality that this paper examines.\par
\subsection{Explaining and Verifying Model Behaviour}
Accuracy on a held-out split does not reveal which pixels a model used, and a recognizer that keys on a recurring background can score well and fail immediately in deployment. Grad-CAM \cite{selvaraju2017grad} and Grad-CAM++ \cite{chattopadhay2018grad} produce class-conditional localization maps that make this visible, but a plausible-looking map is not itself evidence of faithfulness. The deletion and insertion curves of Petsiuk et al. \cite{petsiuk2018rise} test whether the highlighted region is the evidence actually relied on, and the cascading weight-randomization test of Adebayo et al. \cite{adebayo2018sanity} checks that the map depends on the learned parameters rather than the input's edge content. This work uses all four, since each addresses a different way an explanation can mislead.

\section{Related Work}
\label{sec:litreview}
This section reviews prior BdSL recognition research as dataset-oriented, model-oriented, and combined dataset-and-model studies. Table~\ref{tab:litreview} gives the per-study detail along four axes, the data used, the model proposed, the reported result, and the limitation that remains; the discussion below draws out the patterns, so that the gaps addressed by this paper can be traced to specific studies.\par
\begin{sidewaystable*}[p]
\centering
\caption{Summary of the reviewed literature on Bangla Sign Language datasets and recognition models. Studies are grouped by their primary contribution: dataset construction, model design, or both. Reported results are quoted as published and are not directly comparable across rows, since the underlying datasets, class inventories, and evaluation splits differ.}
\label{tab:litreview}
\scriptsize
\setlength{\tabcolsep}{4pt}
\renewcommand{\arraystretch}{1.15}
\begin{tabular}{@{}
>{\raggedright\arraybackslash}p{0.145\textheight}
>{\raggedright\arraybackslash}p{0.225\textheight}
>{\raggedright\arraybackslash}p{0.205\textheight}
>{\raggedright\arraybackslash}p{0.165\textheight}
>{\raggedright\arraybackslash}p{0.205\textheight}
@{}}
\toprule
\textbf{Study} & \textbf{Dataset description} & \textbf{Model description} & \textbf{Reported result} & \textbf{Remaining limitation} \\
\midrule
Hasib et al. (2023) \cite{hasib2023bdsl} & 29,490 images covering 49 classes from 14 adult volunteers using smartphone cameras. & YOLOv4 and deep classification models. & 99.4\% detection with YOLOv4; up to 93\% recognition. & Only 14 volunteers; limited environmental diversity. \\
\addlinespace[2pt]
Hoque et al. (2020) \cite{hoque2020bdsl36} & 1,200 self-collected images and 1,700 external images, augmented with background synthesis. & ResNet50-based classifiers with background augmentation. & Improved visual diversity and robustness. & Visually similar classes cause misclassification; small base set. \\
\addlinespace[2pt]
Das et al. (2023) \cite{das2023hybrid} & Ishara-Lipi and Ishara-Bochon character and digit data. & Deep transfer learning with a random forest classifier. & 91.67\% character accuracy using VGG16 and random forest. & Small dataset; character recognition still weak. \\
\addlinespace[2pt]
Siddique et al. (2023) \cite{siddique2023deep} & Okkhornama dataset with 49 classes and more than 110 images per class. & Edge-device detection models, including YOLOv7 and Faster R-CNN. & 85--97\% detection performance. & Small custom set; mostly teenage volunteers. \\
\addlinespace[2pt]
Rayeed et al. (2023) \cite{rayeed2023bdsl47} & 4,700 RGB images from 10 participants covering 47 classes. & Landmark-based artificial neural network (ANN) baseline. & 97.84\% accuracy with the ANN. & Controlled capture settings; 100 images per class. \\
\addlinespace[2pt]
Jim et al. (2023) \cite{jim2023ku} & 1,500 images from 39 participants covering 30 classes. & Open dataset resource paper. & Benchmark resource for later research. & Only 1,500 images; consonants only. \\
\addlinespace[2pt]
Hasan et al. (2021) \cite{hasan2021shongket} & 5,820 images covering 36 letters and 10 digits. & Classical and deep classifiers, including a CNN. & 95\% accuracy on digits and 91.4\% on letters. & Small scale; grayscale only; no attention mechanisms. \\
\addlinespace[2pt]
Kabir et al. (2025) \cite{kabir2025combining} & BdSL-38 and BDSL-49 datasets. & Max-voting ensemble of five pretrained CNNs. & 96.62\% on BdSL-38 and 99.92\% on BDSL-49. & Very heavy ensemble; source datasets have few participants. \\
\addlinespace[2pt]
Diba et al. (2024) \cite{diba2024explainable} & BdSL47 with landmark-based representations. & Federated learning using FedAvg, FedProx, and FedOpt. & 98.36\% test accuracy with VGG19 and FedAvg. & Depends on landmark preprocessing; structured data only. \\
\addlinespace[2pt]
Sraboni et al. (2024) \cite{sraboni2024real} & BDSL-49 dataset. & Transfer learning models, including MobileNet and Xception. & Xception achieved 97.86\% accuracy. & Signs still misrecognized in real-time use. \\
\addlinespace[2pt]
Raihan et al. (2024) \cite{raihan2024bengali} & Augmented KU-BdSL dataset. & CNN with squeeze-and-excitation blocks and SHapley Additive exPlanations (SHAP) interpretation. & 99.86\% test accuracy. & No benchmarking against other public BdSL datasets. \\
\addlinespace[2pt]
Shams et al. (2024) \cite{shams2024multimodal} & BdSL dataset with 38 classes and 12,581 images. & Multimodal ensemble with three feature streams. & Training accuracies of 99.77\%, 98.11\%, and 99.30\%. & Heavy preprocessing; MediaPipe dependency; static signs. \\
\addlinespace[2pt]
Tapu et al. (2025) \cite{11013937} & BdSL dataset with 38 classes and 12,160 images. & Self-attention CNN with 671,942 parameters. & Accuracy improved from 92.32\% to 93.47\%. & Single dataset; no explainability; limited augmentation. \\
\addlinespace[2pt]
Tasnim et al. (2026) \cite{tasnim2026vision} & BDSL49 dataset. & Fine-tuned DINOv2 ViT-S/14 with 21M parameters. & 99.7\% accuracy and 99.7\% F1-score. & Heavyweight; offline only; no edge deployment test. \\
\addlinespace[2pt]
Billah et al. (2022) \cite{billah2022recognition} & 2,660 images covering 36 letters and 10 digits. & Transfer learning CNNs. & ResNet152V2 achieved 98.5\% accuracy. & Small dataset; heavy reliance on augmentation. \\
\addlinespace[2pt]
Hadiuzzaman et al. (2024) \cite{hadiuzzaman2024baust} & 18,000 images covering 36 alphabets. & Hybrid CNN-LSTM model. & 97.28\% accuracy. & $50\times50$ inputs discard fine hand detail. \\
\addlinespace[2pt]
Islam et al. (2022) \cite{islam2022sign} & 2,340 RGB images with varying backgrounds. & Custom CNN architecture. & 92\% accuracy. & Weak generalization beyond own dataset. \\
\addlinespace[2pt]
Tanvir et al. (2021) \cite{tanvir2021real} & 3,600 images for 36 characters from 12 volunteers. & 2D CNN with adaptive thresholding. & 99.72\% validation accuracy. & Only 12 student signers; controlled setting. \\
\addlinespace[2pt]
Alam et al. (2021) \cite{alam2021two} & 4,600 images for 36 letters and 10 digits. & 2D CNN recognition and translation pipeline. & 99.57\% accuracy. & Limited signer diversity. \\
\addlinespace[2pt]
Emon et al. (2025) \cite{emon2025real} & 9,000 images across 45 classes, augmented to 36,000 images. & Google Teachable Machine models. & 99.98\% accuracy. & Limited model transparency; heavy augmentation. \\
\addlinespace[2pt]
Karim et al. (2025) \cite{karim2025empowering} & 300 images for numerals from 30 participants. & MediaPipe landmarks with a support vector machine (SVM) and LightGBM. & 100\% with SVM and 95.60\% with LightGBM. & Numerals only; extremely small dataset. \\
\addlinespace[2pt]
Podder et al. (2022) \cite{podder2022bangla} & 132,061 images across 87 classes, including 2,300 images for segmentation. & Classification and segmentation pipelines. & 99.99\% accuracy with ResNet18. & No cross-dataset or signer-independent testing. \\
\bottomrule
\end{tabular}
\end{sidewaystable*}
\subsection{Dataset-Oriented Studies}
Dataset development has been a key part of BdSL research, as recognition performance depends heavily on dataset size, class coverage, signer diversity, and background variation. The largest public resource is BDSL49, with 29,490 images over 49 classes of alphabets, numerals, and special characters collected from 14 adult volunteers, released together with detection and recognition benchmarks \cite{hasib2023bdsl}. BdSL36 took a different route, expanding a modest base of self-collected and external images through augmentation and large-scale background synthesis to improve visual diversity and robustness to background variation \cite{hoque2020bdsl36}.\par
The remaining resources are smaller and more specialized. Ishara-Lipi and Ishara-Bochon expose how little usable public BdSL character data exists \cite{das2023hybrid}; Okkhornama supports 49-class detection with more than 110 images per class and targets real-time edge devices \cite{siddique2023deep}; BdSL47 adds landmark-based representations to 4,700 RGB images of 47 classes from 10 participants \cite{rayeed2023bdsl47}; KU-BdSL offers 1,500 images across 30 classes from 39 participants as an open resource paper \cite{jim2023ku}; and Shongket covers 36 letters and 10 digits in 5,820 images, with a CNN reaching 95\% on digits and 91.4\% on letters \cite{hasan2021shongket}. Across this group, the recurring pattern is a small contributor pool, a controlled capture setting, and the absence of any documented linguistic review of the collected images.\par
\subsection{Model-Oriented Studies}
Model-oriented research has pursued accuracy through transfer learning, ensembles, explainable AI, and privacy-preserving frameworks. A max-voting ensemble of five pretrained CNNs (Xception, InceptionV3, DenseNet121, ResNet50, and MobileNetV2) reaches 96.62\% on BdSL-38 and 99.92\% on BDSL-49, showing the effectiveness of ensemble-based decision fusion \cite{kabir2025combining}, while federated training with FedAvg, FedProx, and FedOpt on landmark representations of BdSL47 reports 98.36\% with VGG19 and FedAvg, demonstrating that privacy-preserving learning need not cost recognition performance \cite{diba2024explainable}.\par
Transfer learning is the most widely explored single strategy: among MobileNet, Xception, ResNet50, and InceptionV3 on BDSL-49, Xception attains the best result at 97.86\% \cite{sraboni2024real}, and a CNN with squeeze-and-excitation blocks \cite{hu2018squeeze}, SHAP interpretation, and a Flutter front end reaches 99.86\% on an augmented KU-BdSL \cite{raihan2024bengali}. A multimodal ensemble over spatial, skeletal, and edge-based streams reports training accuracies of 99.77\%, 98.11\%, and 99.30\% on a 38-class corpus \cite{shams2024multimodal}. Closest to the present setting, a self-attention CNN (SA-CNN) of 671,942 parameters improves a plain CNN from 92.32\% to 93.47\% on a 38-class dataset, showing that attention pays off even within compact architectures \cite{11013937}; at the opposite extreme, a fine-tuned DINOv2 vision transformer (ViT-S/14) reaches 99.7\% accuracy and F1-score on BDSL49 \cite{tasnim2026vision}. The accuracy ceiling in this group is high, but it is reached either by ensembling several heavyweight backbones or by fine-tuning a 21M-parameter transformer, and no study in the group reports the compute, memory, or latency that such a system would require on a mobile device.\par
\subsection{Combined Dataset-and-Model Studies}
A large portion of BdSL research follows a combined strategy in which authors first construct or extend a dataset and then validate a proposed model on that same dataset. Examples span a 2,660-image letter-and-digit corpus on which ResNet152V2 reaches 98.5\% \cite{billah2022recognition}; BAUST Lipi, an 18,000-image set of 36 alphabets from 15 volunteers paired with a hybrid CNN-LSTM at 97.28\% \cite{hadiuzzaman2024baust}; and a 2,340-sample RGB set with varied backgrounds classified at 92\% by a custom CNN \cite{islam2022sign}.\par
Real-time recognition and translation prototypes follow the same pattern on self-curated data: 3,600 images from 12 volunteers with a 2D CNN and adaptive thresholding at 99.72\% \cite{tanvir2021real}, a 4,600-image extension covering 36 letters and 10 digits at 99.57\% \cite{alam2021two}, and a 9,000-image set augmented to 36,000 and classified by Google Teachable Machine models at 99.98\% \cite{emon2025real}. More specialized directions include numeral recognition from 300 images of 30 participants using MediaPipe landmarks, where SVM reaches 100\% and LightGBM 95.60\% \cite{karim2025empowering}, and the large BdSL-D1500 corpus of roughly 132,061 images across 87 classes with the companion BdSLHD-2300 segmentation set, on which ResNet18 attains 99.99\% \cite{podder2022bangla}. The trend toward practical prototypes is clear, but the near-perfect accuracies are obtained on splits drawn from the same small volunteer pool that produced the training images.\par
Several research gaps emerge. Nearly all public BdSL datasets were captured from small pools of hearing volunteers (10 to 42 people) in controlled environments without expert verification, so models trained on them may learn volunteer-specific articulations rather than authentic BdSL. The strongest reported results rely on heavyweight pretrained backbones such as ResNet152V2 and the 21M-parameter DINOv2, or on multi-model ensembles \cite{kabir2025combining,tasnim2026vision}, which suit on-device assistive applications poorly, and none reports compute cost, quantized size, or measured on-device latency. The few lightweight designs, such as SA-CNN \cite{11013937}, were validated on a single dataset only. Cross-dataset evaluation and component-level ablation are almost entirely absent, leaving the contribution of individual architectural components unquantified \cite{podder2022bangla,islam2022sign}, and signer-independent evaluation is effectively unreported, so the optimism introduced by signer overlap in stratified splits is unknown across the field. Our proposed dataset, model, and evaluation protocol address all of these gaps.
\section{The RSBdSL38 Dataset}
\label{sec:dataset}
\begin{figure*}[!t]
    \centering
    \includegraphics[width=0.8\textwidth]{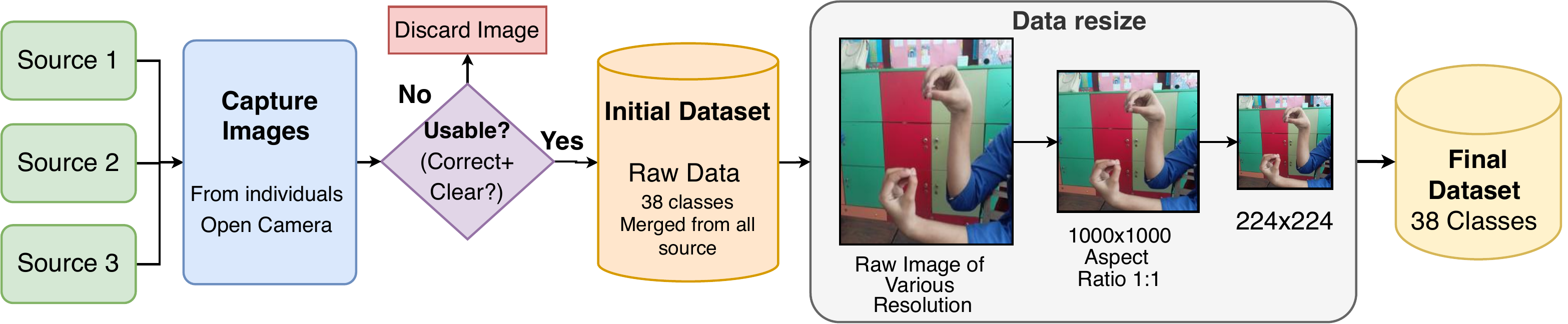}
    \caption{Creation pipeline of the RSBdSL38 dataset: participant recruitment at three special-needs schools in three districts of Bangladesh together with additional adult volunteers, smartphone capture under uncontrolled backgrounds, a manual quality-screening pass by the research team followed by expert validation of the class labels, preprocessing to $224\times224$ RGB, and class-stratified partitioning into training, validation, and test sets.}
    \label{fig:data-creation}
\end{figure*}

Existing BdSL datasets are predominantly collected from hearing volunteers in controlled laboratory conditions, and most provide no verification by a sign language expert (Section~\ref{sec:litreview}). RSBdSL38 (Real-Signer Bangla Sign Language, 38 classes)\cite{ahmed2026rsbdsl38data} was therefore designed around three principles: \emph{authentic signers}, \emph{expert verification}, and \emph{environmental diversity}. Fig.~\ref{fig:data-creation} illustrates the complete creation pipeline.\par
\textbf{(i) Participants and Collection Protocol.} Images were collected from real signers at three special-needs schools in three different districts of Bangladesh, where BdSL is the daily medium of communication, supplemented by additional volunteers to broaden hand-shape and skin-tone diversity. In total, \textbf{36 participants} contributed (17 male, 19 female) in two age bands: 21 aged 8--15 years, the deaf and hard-of-hearing pupils recruited at the schools, and 15 aged 20--28 years, the adult volunteers (Table~\ref{tab:demographics}). Mixing native child signers with adult contributors means the retained hand shapes reflect both fluent day-to-day articulation and a wider range of hand morphology and skin tone. All signs were captured with smartphone cameras framed on the hand region and articulated naturally, without constraining hand orientation or requiring uniform backgrounds, so the corpus contains the classroom walls, furniture, floor tiling, and variable illumination of the recording sites. Informed consent was obtained from all participants or their guardians.\par
\begin{figure*}[!t]
    \centering
    \includegraphics[width=0.9\textwidth]{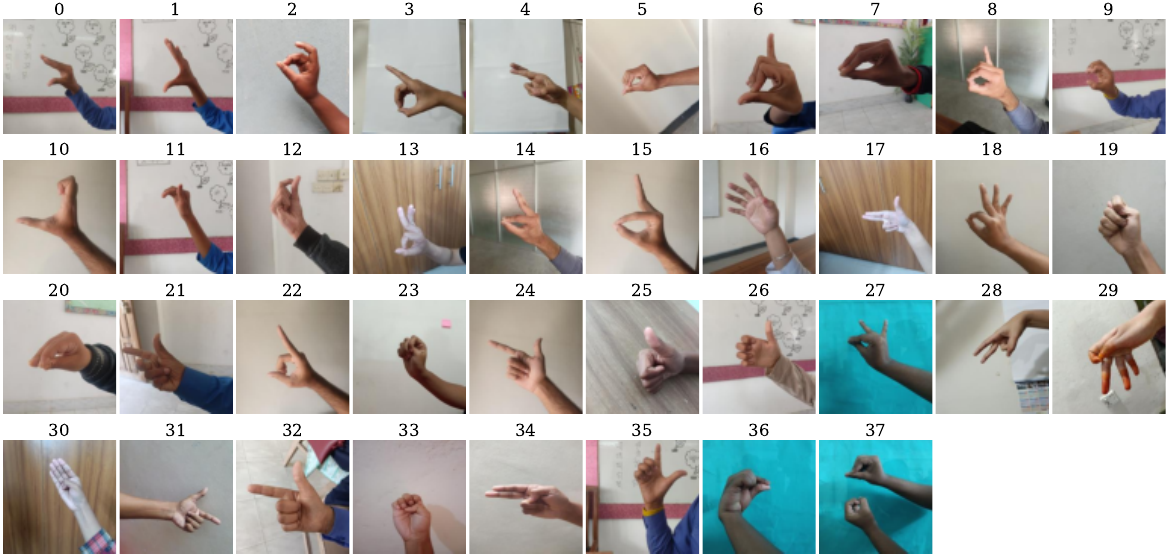}
    \caption{Representative samples of the 38 canonical BdSL hand signs in RSBdSL38, one randomly drawn image per class. The panels illustrate the uncontrolled capture conditions of the corpus, including classroom walls, furniture, tiled floors, and variable illumination, as well as the fine articulatory differences that separate several of the classes.}
    \label{fig:sign-grid}
\end{figure*}

\begin{table}[!t]
  \centering
  \caption{Participant demographics of RSBdSL38. The cohort combines deaf and hard-of-hearing pupils recruited at three special-needs schools with adult volunteers recruited to broaden hand morphology and skin-tone coverage.}
  \label{tab:demographics}
  \setlength{\tabcolsep}{6pt}
  \footnotesize
  \begin{tabular}{@{}l l r@{}}
    \toprule
    \textbf{Attribute} & \textbf{Group} & \textbf{Count} \\
    \midrule
    \multirow{2}{*}{Gender} & Male   & 17 \\
                            & Female & 19 \\
    \midrule
    \multirow{2}{*}{Age (years)} & 8--15 (school signers) & 21 \\
                                 & 20--28 (adult volunteers) & 15 \\
    \midrule
    \multicolumn{2}{@{}l}{\textbf{Total participants}} & \textbf{36} \\
    \bottomrule
  \end{tabular}
\end{table}
\textbf{(ii) Class Inventory.} RSBdSL38 covers the complete canonical BdSL manual alphabet: 38 sign classes jointly representing the 51 letters of the written Bangla alphabet \cite{nihal2021bangla}, which makes it directly comparable with the 38-class benchmarks of recent literature \cite{kabir2025combining,shams2024multimodal,11013937} while providing complete alphabet coverage. Of the 38 signs, 9 are vowels (\emph{shoroborno}) and the remaining 29 consonants and diacritic modifiers (\emph{banjonborno}), with written letters that share an identical manual form grouped under a single sign. Table~\ref{tab:sign-map} gives the correspondence between class labels and Bangla letters, and Fig.~\ref{fig:sign-grid} shows the 38 signs.\par
\begin{table*}[!t]
  \centering
  \caption{Mapping between the 38 RSBdSL38 class labels and the Bangla letters they represent. A single manual sign covers several written letters wherever those letters share an identical canonical hand form, which is why 38 signs span the 51 letters of the Bangla alphabet.}
  \label{tab:sign-map}
  \setlength{\tabcolsep}{5pt}
  \renewcommand{\arraystretch}{1.0}
  \footnotesize
  \begin{tabular}{@{}c l @{\hspace{2em}} c l @{\hspace{2em}} c l @{\hspace{2em}} c l @{\hspace{2em}} c l@{}}
    \toprule
    \textbf{Label} & \textbf{Letter} & \textbf{Label} & \textbf{Letter} & \textbf{Label} & \textbf{Letter} & \textbf{Label} & \textbf{Letter} & \textbf{Label} & \textbf{Letter} \\
    \midrule
    0 & \bn{অ} & 8 & \bn{ঔ} & 16 & \bn{জ/য} & 24 & \bn{ত/ৎ} & 32 & \bn{ল} \\
    1 & \bn{আ} & 9 & \bn{ক} & 17 & \bn{ঝ} & 25 & \bn{থ} & 33 & \bn{শ/স/ষ} \\
    2 & \bn{ই/ঈ} & 10 & \bn{খ/ক্ষ} & 18 & \bn{ঞ} & 26 & \bn{দ} & 34 & \bn{হ} \\
    3 & \bn{উ/ঊ} & 11 & \bn{গ} & 19 & \bn{ট} & 27 & \bn{ধ} & 35 & \bn{ঁ} \\
    4 & \bn{ঋ/ৠ/ৃ} & 12 & \bn{ঘ} & 20 & \bn{ঠ} & 28 & \bn{প} & 36 & \bn{ং} \\
    5 & \bn{এ} & 13 & \bn{ঙ} & 21 & \bn{ড} & 29 & \bn{ফ} & 37 & \bn{ঃ} \\
    6 & \bn{ঐ} & 14 & \bn{চ} & 22 & \bn{ঢ} & 30 & \bn{ব/ভ} &  &  \\
    7 & \bn{ও} & 15 & \bn{ছ} & 23 & \bn{ন} & 31 & \bn{ম} &  &  \\
    \bottomrule
  \end{tabular}
\end{table*}
\textbf{(iii) Quality Screening.} Prior to expert validation, the research team performed a manual screening pass over the collected images to remove samples unsuitable for recognition, discarding those with severe blur, a truncated or partially out-of-frame hand, or otherwise poor framing in which the hand shape could not be clearly discerned. This step filtered low-quality captures on purely visual grounds and is independent of the linguistic label check described next. The retained images were standardised to $224\times224$ RGB for model input, with the training-time rescaling and augmentation detailed in Section~\ref{sec:methodology}.\par
\textbf{(iv) Expert Validation.} After the research team assigned an initial class label to every image, the complete dataset was independently reviewed by an experienced senior Bangla Sign Language teacher at one of the participating schools, who instructs deaf and hard-of-hearing students in BdSL as part of her regular duties. The images were organised into their 38 class folders and examined class by class on a computer: for each class, the expert confirmed whether the grouped images corresponded to the correct canonical BdSL sign for that label. Images identified as mislabelled or incorrectly articulated were flagged, and the corresponding corrections were applied by the research team. This independent expert review, in place of the volunteer self-labelling relied upon by prior BdSL corpora, yields the validated corpus of \textbf{10{,}874 images}, distributed approximately uniformly across the 38 classes ($\approx$286 per class) and partitioned by class-stratified sampling into training, validation, and test sets of 8{,}794 / 977 / 1{,}103 images (80.9\% / 9.0\% / 10.1\%). The same frozen partition is used for the proposed model and for every baseline in Section~\ref{sec:results}. A signed validation statement from the expert is retained by the authors and available on request.\par
\begin{figure}[!t]
    \centering
    \includegraphics[width=\columnwidth]{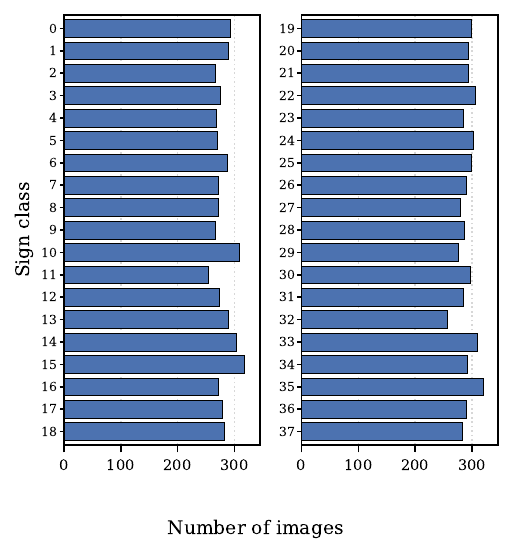}
    \caption{Per-class image distribution of RSBdSL38 across the 38 sign classes. Counts are near-uniform (mean $286.2$, standard deviation $15.5$, maximum-to-minimum imbalance ratio $1.25$), so weighted and macro-averaged metrics are effectively equivalent on this benchmark.}
    \label{fig:distribution}
\end{figure}

\textbf{(v) Class Balance.} The dataset is approximately balanced. Fig.~\ref{fig:distribution} shows the per-class image counts: mean $286.2$ per class, standard deviation $15.5$, and a maximum-to-minimum imbalance ratio of only $1.25$ (largest class $=320$, smallest $=255$). This near-uniform distribution avoids the majority-class bias that can otherwise inflate weighted metrics, so no single class dominates the metrics reported in Section~\ref{sec:results} and the weighted and macro averages agree to within $0.02$ percentage points.\par
\begin{table*}[!t]
  \centering
  \caption{RSBdSL38 compared with existing public BdSL datasets. Several existing corpora contain more raw images, but RSBdSL38 is the only one that combines full 38-class alphabet coverage, genuine signers recruited at multiple geographic sites, uncontrolled capture conditions, and complete expert validation.}
  \label{tab:datasetcomp}
  \setlength{\tabcolsep}{6pt}
  \footnotesize
  \begin{tabular}{@{}l c r l c@{}}
    \toprule
    \textbf{Dataset} & \textbf{Classes} & \textbf{Images} & \textbf{Contributors} & \textbf{Expert verified} \\
    \midrule
    BDSL49 \cite{hasib2023bdsl}            & 49 & 29,490 & 14 volunteers        & No \\
    BdSL36 \cite{hoque2020bdsl36}          & 36 &  2,712 & Volunteers           & Yes$^{\dagger}$ \\
    KU-BdSL \cite{jim2023ku}               & 30 &  1,500 & 39 volunteers        & No \\
    BdSL47 \cite{rayeed2023bdsl47}         & 47 &  4,700 & 10 volunteers        & No \\
    Shongket \cite{hasan2021shongket}      & 46 &  5,820 & Volunteers           & No \\
    BAUST Lipi \cite{hadiuzzaman2024baust} & 36 & 18,000 & 15 volunteers        & No \\
    BdSL-38 \cite{kabir2025combining}      & 38 & 12,581$^{\ddagger}$ & Volunteers & No \\
    \midrule
    \textbf{RSBdSL38 (ours)} & \textbf{38} & \textbf{10,874} & \textbf{36 real signers and volunteers} & \textbf{Yes} \\
    \bottomrule
  \end{tabular}
  \\[3pt]
  {\footnotesize $^{\dagger}$BdSL36 images were collected at a deaf school and individually filtered by BdSL signer experts \cite{hoque2020bdsl36}; RSBdSL38 differs in combining real signers from multiple sites, full 38-class alphabet coverage, and uncontrolled capture with complete expert validation. $^{\ddagger}$Usable image counts reported for this release vary slightly across studies (12{,}160 in \cite{11013937}); we quote the 12{,}581 images of the public distribution, which is the set used for the zero-shot transfer experiment of Section~\ref{sec:zeroshot}.}
\end{table*}
Table~\ref{tab:datasetcomp} contrasts RSBdSL38 with the public datasets used in this study. Several contain more raw images, but RSBdSL38 is unique in combining full-alphabet coverage, genuine BdSL signers from multiple geographic sites, uncontrolled capture conditions, and complete expert validation, the properties most directly tied to real-world generalization (RQ4). Because participant identity is recorded for every image, it additionally supports the signer-independent protocol of Section~\ref{sec:signerindep}, which none of the compared datasets documents.
\section{Methodology}
\label{sec:methodology}
This section presents the technical details of the proposed recognizer: its operators, the building blocks that constitute a stage, the composition of the complete network, its parameter and compute budget, and the training procedure. Every symbol used in this paper are defined and summarized in ~\ref{app:summaryofnotations}; the layer-level architecture diagram, the individual block diagrams, and the stage-wise configuration table are given in ~\ref{app:architecture}.

\subsection{Design Principles}
\begin{figure*}[!t]
    \centering
    \includegraphics[width=0.85\textwidth]{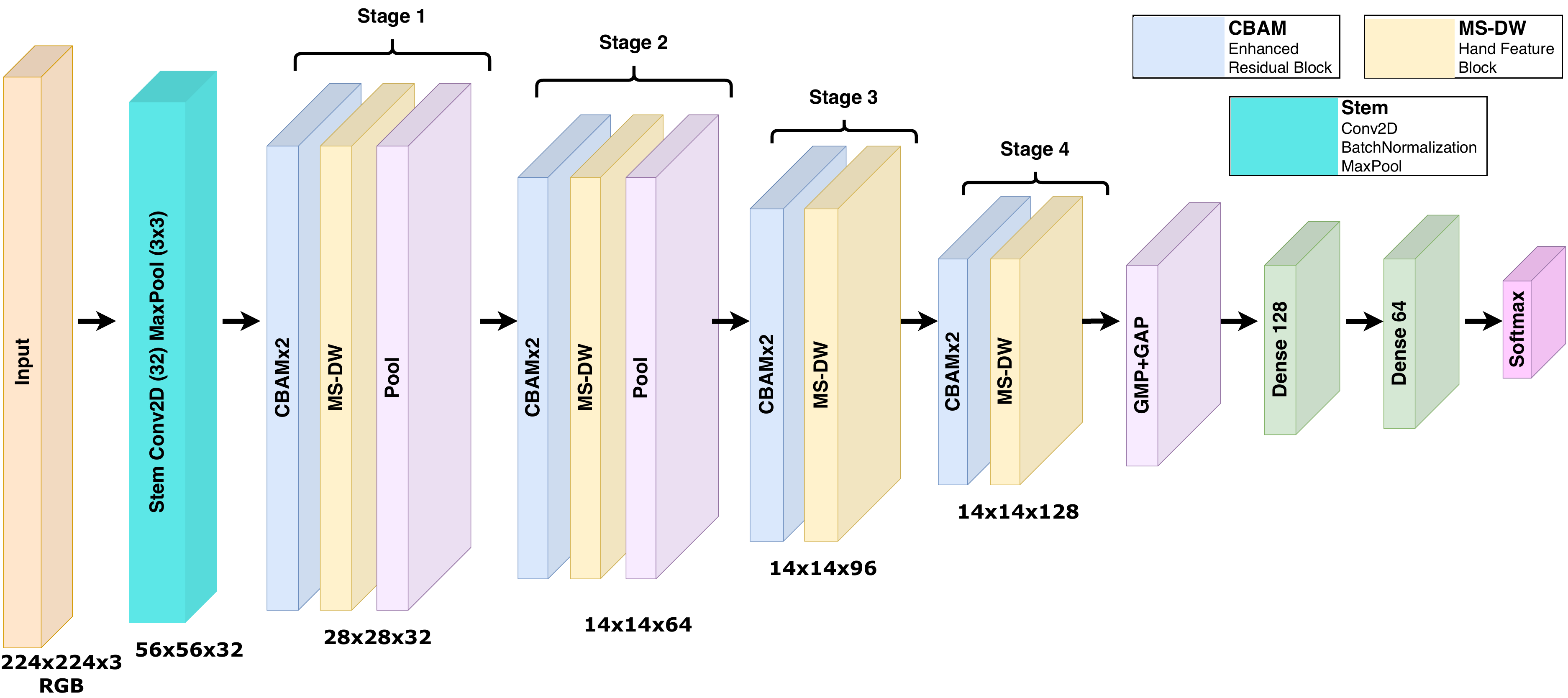}
    \caption{Overall architecture of the proposed lightweight recognizer. A strided convolutional stem reduces the $224\times224$ input by a factor of four and is followed by four stages of width 32, 64, 96, and 128. Each stage contains two grouped bottleneck residual attention blocks followed by a multi-scale depthwise hand-feature block; stages 1 and 2 end in $2\times2$ max pooling, whereas stages 3 and 4 preserve the $14\times14$ resolution. A dual-pooling head concatenates global average and global max descriptors and passes them through two fully connected layers to a 38-way softmax. The layer-level diagram and the internal structure of each block are given in Appendix~\ref{app:architecture}.}
    \label{fig:model}
\end{figure*}

The overall architecture is shown in Fig.~\ref{fig:model}: the network maps a $224 \times 224 \times 3$ RGB image to a probability distribution over the $K = 38$ classes using only 298{,}470 parameters. Four principles govern the design. First, \emph{bottleneck factorization with grouped convolutions} \cite{he2016deep,xie2017aggregated} keeps the spatial layers small enough that all four stages fit within 0.30M weights. Second, \emph{convolutional block attention} \cite{woo2018cbam} directs that limited capacity towards the signing hand rather than the background, which matters because RSBdSL38 is captured against uncontrolled scenes. Third, a dedicated \emph{multi-scale depthwise hand-feature block} \cite{chollet2017xception} captures finger-level and palm-level structure simultaneously, the two characteristic scales at which BdSL signs differ. Fourth, \emph{smooth activations with strong regularization} \cite{ramachandran2017swish,ioffe2015batch,tompson2015efficient,srivastava2014dropout} compensate for a training set of fewer than nine thousand images.\par
\subsection{Operators and Preliminaries}
Three operators recur throughout the network and are defined here once. A \emph{grouped convolution} with $g$ groups partitions an input $\mathbf{X}\in\mathbb{R}^{H\times W\times C_{\mathrm{in}}}$ into $g$ disjoint channel blocks $\mathbf{X}^{(1)},\dots,\mathbf{X}^{(g)}$ of $C_{\mathrm{in}}/g$ channels each, convolves every block with its own kernel, and concatenates the results:
\begin{equation}
\mathbf{W} \ast_g \mathbf{X} = \left[\, \mathbf{W}^{(1)} \ast \mathbf{X}^{(1)} \;\|\; \cdots \;\|\; \mathbf{W}^{(g)} \ast \mathbf{X}^{(g)} \,\right],
\label{eq:groupconv}
\end{equation}
where $\mathbf{W}^{(j)} \in \mathbb{R}^{k \times k \times (C_{\mathrm{in}}/g) \times (C_{\mathrm{out}}/g)}$. The kernel cost therefore falls from $k^2 C_{\mathrm{in}} C_{\mathrm{out}}$ to $k^2 C_{\mathrm{in}} C_{\mathrm{out}} / g$ \cite{krizhevsky2012imagenet,xie2017aggregated}. A \emph{depthwise convolution} is the limiting case $g = C_{\mathrm{in}} = C_{\mathrm{out}}$, in which every channel receives its own $k \times k$ kernel and no cross-channel mixing occurs \cite{chollet2017xception}:
\begin{equation}
\left[\mathrm{DW}^{k\times k}(\mathbf{X})\right]_{u,v,c} = \sum_{a=1}^{k}\sum_{b=1}^{k} \mathbf{W}_{a,b,c}\, \mathbf{X}_{u+a,\,v+b,\,c},
\label{eq:dwconv}
\end{equation}
at a cost of $k^2 C$ weights, independent of the number of output channels. \emph{Batch normalization} \cite{ioffe2015batch} standardizes each channel over the mini-batch and restores scale and shift through learned parameters $\gamma$ and $\beta$:
\begin{equation}
\mathrm{BN}(x) = \gamma \cdot \frac{x - \mathbb{E}_B[x]}{\sqrt{\mathrm{Var}_B[x] + \epsilon}} + \beta .
\label{eq:bn}
\end{equation}
At inference the batch statistics are replaced by exponential running estimates, which constitute the network's $6{,}208$ non-trainable parameters. The Swish (SiLU) activation is used throughout,
\begin{equation}
\delta(x) = x \cdot \sigma(x) = \frac{x}{1 + e^{-x}},
\label{eq:swish}
\end{equation}
which is smooth, non-monotonic, and preserves small negative activations; Section~\ref{sec:ablationB} shows this choice worth $+0.99$ percentage points relative to the Rectified Linear Unit (ReLU).\par
\subsection{Model Architecture}
\textbf{(i) Stem.} Given an input image $\mathbf{I} \in \mathbb{R}^{224 \times 224 \times 3}$, the stem $\mathcal{S}_0$ applies a strided $5 \times 5$ convolution with 32 filters followed by batch normalization, Swish, and overlapping max pooling:
\begin{equation}
\mathbf{X}_0 = \mathcal{S}_0(\mathbf{I}) = \mathrm{MaxPool}_{3,\,s=2}\!\left( \delta\!\left( \mathrm{BN}\!\left( \mathbf{W}_{\mathrm{stem}} \ast_{s=2} \mathbf{I} \right) \right) \right),
\label{eq:stem}
\end{equation}
which reduces the resolution by a factor of four to $56 \times 56 \times 32$ using $2{,}432$ convolutional weights (the $2{,}560$ reported for the stem in Fig.~\ref{fig:model_arc_main} additionally counts its $128$ batch-normalization parameters). A single $5\times5$ kernel at stride two is preferred to a stack of small kernels because the earliest layer operates at the highest resolution, where compute is most expensive.\par
\textbf{(ii) Grouped Bottleneck Residual Attention Block.} Each of the four stages $i \in \{1,2,3,4\}$, with channel widths $C_i \in \{32, 64, 96, 128\}$, contains two residual blocks $\mathcal{B}^{(1)}_i$ and $\mathcal{B}^{(2)}_i$. Let $\mathbf{X} \in \mathbb{R}^{H \times W \times C}$ be the block input with $C = C_i$. A bottleneck path compresses the channels to $C/2$, processes them with a grouped $3\times3$ convolution with $g_i \in \{2,4,8,16\}$ groups, and re-expands to $C$:
\begin{align}
\mathbf{U} &= \delta\!\left( \mathrm{BN}\!\left( \mathbf{W}_{c}^{1\times1} \ast \mathbf{X} \right) \right) \in \mathbb{R}^{H\times W\times \frac{C}{2}}, \label{eq:bneck1}\\
\mathbf{V} &= \delta\!\left( \mathrm{BN}\!\left( \mathbf{W}_{g}^{3\times3} \ast_{g_i} \mathbf{U} \right) \right) \in \mathbb{R}^{H\times W\times \frac{C}{2}}, \label{eq:bneck2}\\
\mathbf{Y} &= \mathrm{BN}\!\left( \mathbf{W}_{e}^{1\times1} \ast \mathbf{V} \right) \in \mathbb{R}^{H\times W\times C}. \label{eq:bneck3}
\end{align}
The compression halves the width seen by the $3\times3$ layer and the grouping divides its remaining cost by $g_i$, so the spatial convolution costs $9C^2/(4g_i)$ weights instead of the $9C^2$ of a dense $3\times3$ layer of the same width. The group count scales with the stage width, from $g_1 = 2$ to $g_4 = 16$, keeping the per-group channel count close to constant across depth while the widening stages remain affordable.\par
\textbf{(iii) Channel Attention.} Following the Convolutional Block Attention Module (CBAM) \cite{woo2018cbam}, the feature $\mathbf{Y}$ is first recalibrated along the channel axis. Global average and global max pooling produce two channel descriptors that are passed through a \emph{shared} two-layer bottleneck multi-layer perceptron (MLP) with reduction ratio $r = 16$, summed, and squashed:
\begin{equation}
\mathbf{M}_c = \sigma\!\Big( \mathbf{W}_{a1} \, \phi\big( \mathbf{W}_{a0} \, \mathrm{GAP}(\mathbf{Y}) \big) + \mathbf{W}_{a1} \, \phi\big( \mathbf{W}_{a0} \, \mathrm{GMP}(\mathbf{Y}) \big) \Big),
\label{eq:chatt}
\end{equation}
where $\mathbf{W}_{a0} \in \mathbb{R}^{\frac{C}{r} \times C}$ and $\mathbf{W}_{a1} \in \mathbb{R}^{C \times \frac{C}{r}}$ are bias-free and shared between the two pooling paths, so the module costs only $2C^2/r$ weights. The average descriptor summarizes how strongly a channel responds overall and the max descriptor how strongly it responds at its most active location; sharing the MLP forces both onto the same scale. The channel-refined feature is $\mathbf{Y}' = \mathbf{M}_c \otimes \mathbf{Y}$, where $\otimes$ broadcasts the $C$-dimensional gate over all spatial positions.\par
\textbf{(iv) Spatial Attention.} A single-channel spatial saliency map is then computed with a bias-free $7 \times 7$ convolution and sigmoid gating,
\begin{equation}
\mathbf{M}_s = \sigma\!\left( f^{7\times7}\!\left( \mathbf{Y}' \right) \right) \in \mathbb{R}^{H\times W\times 1}, \qquad
\mathbf{Y}'' = \mathbf{M}_s \otimes \mathbf{Y}',
\label{eq:spatt}
\end{equation}
which suppresses background regions and concentrates the representation on the signing hand. The large $7\times7$ support is affordable because the map has a single output channel, costing $49C$ weights, and lets one gate decision see a neighbourhood wide enough to contain a whole finger group. The block output is formed by a residual addition and Swish,
\begin{equation}
\mathbf{Z} = \delta\!\left( \mathbf{Y}'' + \mathcal{P}(\mathbf{X}) \right),
\label{eq:residual}
\end{equation}
where the shortcut $\mathcal{P}$ is a $1\times1$ convolution with batch normalization in the first block of each stage, where the width changes, and the identity in the second \cite{he2016deep}. Because the attention gates are multiplicative and bounded in $(0,1)$, the identity path is what prevents attenuation across eight successive gated blocks.\par
\textbf{(v) Hand-Feature Block.} BdSL signs are distinguished at two characteristic scales: individual finger configuration (fine) and overall hand pose (coarse). To capture both cheaply, each stage ends with a multi-scale block $\mathcal{H}_i$ built from depthwise convolutions. Given the stage feature $\mathbf{Z}$, a $1\times1$ projection is followed by two parallel depthwise branches with $3\times3$ and $5\times5$ kernels, whose outputs are concatenated, fused, and regularized:
\begin{align}
\mathbf{H}   &= \delta\!\left( \mathrm{BN}\!\left( \mathbf{W}_{p}^{1\times1} \ast \mathbf{Z} \right) \right), \label{eq:hfb1}\\
\mathbf{H}_3 &= \delta\!\left( \mathrm{BN}\!\left( \mathrm{DW}^{3\times3}(\mathbf{H}) \right) \right), \quad
\mathbf{H}_5 = \delta\!\left( \mathrm{BN}\!\left( \mathrm{DW}^{5\times5}(\mathbf{H}) \right) \right), \label{eq:hfb2}\\
\mathbf{Z}' &= \mathrm{SD}_{\rho_i}\!\Big( \delta\!\Big( \mathrm{BN}\!\Big( \mathbf{W}_{f}^{1\times1} \ast \left[ \mathbf{H}_3 \,\|\, \mathbf{H}_5 \right] \Big) \Big) \Big), \label{eq:hfb3}
\end{align}
with stage-wise increasing spatial dropout rates $\rho_i \in \{0.05, 0.10, 0.15, 0.20\}$ \cite{tompson2015efficient}. Spatial dropout removes entire feature maps rather than individual activations, the appropriate form of noise here because neighbouring activations within one map are strongly correlated and element-wise dropout would leave the channel largely intact. The two depthwise branches contribute only $34C_i$ weights, so the multi-scale capacity is nearly free; the $1\times1$ projection and fusion account for the block's remaining $3C_i^2$ weights.\par
\textbf{(vi) Stage Composition.} A stage applies its two residual blocks, then its hand-feature block, then a transition $\pi_i$:
\begin{equation}
\mathbf{Z}'_i = \mathcal{S}_i\!\left( \mathbf{Z}'_{i-1} \right) = \Big( \pi_i \circ \mathcal{H}_i \circ \mathcal{B}^{(2)}_i \circ \mathcal{B}^{(1)}_i \Big)\!\left( \mathbf{Z}'_{i-1} \right), \quad \mathbf{Z}'_0 = \mathbf{X}_0,
\label{eq:stage}
\end{equation}
where $\pi_i = \mathrm{MaxPool}_{2,\,s=2}$ for $i \in \{1,2\}$ and $\pi_i = \mathrm{id}$ for $i \in \{3,4\}$. The four stages therefore operate on feature maps of $56^2 \times 32$, $28^2 \times 64$, $14^2 \times 96$, and $14^2 \times 128$, and the resolution is deliberately held at $14 \times 14$ through the last two stages: further downsampling would collapse the finger detail on which several classes depend, so the widening channel count supplies the additional capacity instead.\par
\textbf{(vii) Dual-Pooling Classification Head.} Global max pooling preserves the single strongest response, useful when a sign's identity hinges on one localized finger configuration, whereas global average pooling summarizes the overall activation pattern and is more robust to noise. The head $\mathcal{C}$ concatenates both rather than choosing between them:
\begin{equation}
\mathbf{v} = \left[ \mathrm{GMP}(\mathbf{Z}'_4) \,\|\, \mathrm{GAP}(\mathbf{Z}'_4) \right] \in \mathbb{R}^{256}.
\label{eq:dualpool}
\end{equation}
The vector $\mathbf{v}$ is processed by two fully connected layers with Swish, batch normalization, and dropout ($p = 0.3$),
\begin{align}
\mathbf{h}_1 &= \mathrm{BN}\!\left( \delta\!\left( \mathbf{W}_{fc1} \, \mathrm{Drop}_{p}(\mathbf{v}) + \mathbf{b}_1 \right) \right) \in \mathbb{R}^{128}, \label{eq:fc1}\\
\mathbf{h}_2 &= \mathrm{BN}\!\left( \delta\!\left( \mathbf{W}_{fc2} \, \mathrm{Drop}_{p}(\mathbf{h}_1) + \mathbf{b}_2 \right) \right) \in \mathbb{R}^{64}, \label{eq:fc2}
\end{align}
and the posterior over the $K = 38$ classes is produced by softmax:
\begin{equation}
\hat{y}_k = \frac{\exp\!\left( \mathbf{w}_k^{\top} \mathbf{h}_2 + b_k \right)}{\sum_{j=1}^{K} \exp\!\left( \mathbf{w}_j^{\top} \mathbf{h}_2 + b_j \right)}, \qquad k = 1, \ldots, K.
\label{eq:softmax}
\end{equation}
The complete network is thus the composition
\begin{equation}
\hat{\mathbf{y}} = \mathcal{F}(\mathbf{I}) = \Big( \mathcal{C} \circ \mathcal{S}_4 \circ \mathcal{S}_3 \circ \mathcal{S}_2 \circ \mathcal{S}_1 \circ \mathcal{S}_0 \Big)(\mathbf{I}).
\label{eq:network}
\end{equation}
\subsection{Parameter and Compute Budget}
Counting only kernel weights, a residual block and a hand-feature block of width $C_i$ cost
\begin{align}
\left| \mathcal{B}_i \right| &= \underbrace{\tfrac{1}{2}C_i^2}_{\text{compress}} + \underbrace{\tfrac{9C_i^2}{4 g_i}}_{\text{grouped } 3\times3} + \underbrace{\tfrac{1}{2}C_i^2}_{\text{expand}} + \underbrace{\tfrac{2C_i^2}{r}}_{\text{channel attn.}} + \underbrace{49 C_i}_{\text{spatial attn.}}, \label{eq:paramB}\\
\left| \mathcal{H}_i \right| &= \underbrace{C_i^2}_{\text{project}} + \underbrace{34 C_i}_{\text{DW } 3\times3,\,5\times5} + \underbrace{2C_i^2}_{\text{fuse}} . \label{eq:paramH}
\end{align}
For the widest stage ($C_4 = 128$, $g_4 = 16$) a complete residual block occupies $27{,}776$ parameters including biases and normalization, roughly $5.3\times$ fewer than the $147{,}584$ a dense $3\times3$ convolution of the same width would need. Over the stem, the eight residual blocks, the four hand-feature blocks, and the head, the network contains $298{,}470$ parameters, $292{,}262$ trainable and $6{,}208$ non-trainable batch-normalization statistics, occupying $1.14$~MB in single precision: roughly $8.5\times$ fewer than the smallest pretrained baseline of Section~\ref{sec:results} and $68\times$ fewer than the largest. The inference cost follows from summing the per-layer multiply--accumulate (MAC) counts,
\begin{equation}
\mathrm{MACs} = \sum_{l} H_l W_l \, k_l^2 \, \frac{C^{\mathrm{in}}_l C^{\mathrm{out}}_l}{g_l} = 132.7\ \text{M},
\label{eq:macs}
\end{equation}
at a $224\times224$ input, lower than every baseline in the comparison suite of Section~\ref{sec:efficiency}. Table~\ref{tab:arch} of Appendix~\ref{app:architecture} summarizes the stage-wise configuration.\par
\subsection{Training Objective and Optimization}
The network is trained end-to-end by minimizing the $\ell_2$-regularized categorical cross-entropy
\begin{equation}
\mathcal{L} = -\frac{1}{N} \sum_{n=1}^{N} \sum_{k=1}^{K} y_{nk} \log \hat{y}_{nk} \; + \; \lambda \sum_{l} \left\lVert \mathbf{W}_l \right\rVert_2^2,
\label{eq:loss}
\end{equation}
with $\lambda = 10^{-5}$ applied to the stem kernel, the three bottleneck kernels and the shortcut projection of every residual block, and the two hidden fully connected layers. Optimization uses stochastic gradient descent (SGD) with Nesterov momentum \cite{sutskever2013importance},
\begin{align}
\mathbf{m}_{t+1} &= \mu \, \mathbf{m}_t - \eta_t \, \nabla_{\!\boldsymbol{\theta}} \mathcal{L}\!\left( \boldsymbol{\theta}_t + \mu \, \mathbf{m}_t \right), \label{eq:sgd1}\\
\boldsymbol{\theta}_{t+1} &= \boldsymbol{\theta}_t + \mathbf{m}_{t+1}, \label{eq:sgd2}
\end{align}
with $\mu = 0.9$ and initial learning rate $\eta_0 = 10^{-2}$. Whenever the training loss fails to improve for five consecutive epochs the learning rate is halved, subject to a floor $\eta_{\min} = 10^{-8}$,
\begin{equation}
\eta_{t+1} = \max\!\left( \tfrac{1}{2} \eta_t, \; \eta_{\min} \right),
\label{eq:lrschedule}
\end{equation}
and training stops early if the training loss has not improved for thirty consecutive epochs, restoring the best weights, with a maximum budget of 300 epochs. Three checkpoints are retained throughout (best validation loss, best validation accuracy, and best training loss), and all reported results use the best-validation-accuracy checkpoint. The batch size is $B = 32$, pixel intensities are rescaled to $[0,1]$ for all three partitions, and all convolutional and dense kernels use He normal initialization \cite{he2016deep}.\par
\subsection{Data Augmentation}
To improve robustness to viewpoint and illumination variation, each training image is passed through a stochastic geometric operator followed by a stochastic photometric operator,
\begin{equation}
\tilde{\mathbf{I}} = \left( \tau_{\mathrm{pho}} \circ \tau_{\mathrm{geo}} \right)\!\left( \mathbf{I} \right),
\label{eq:augment}
\end{equation}
where $\tau_{\mathrm{geo}}$ composes a random rotation of up to $\pm15^{\circ}$, a random zoom of $\pm10\%$, and a random translation of $\pm10\%$ along each axis, and $\tau_{\mathrm{pho}}$ composes random brightness ($\pm0.08$), contrast ($\pm10\%$), saturation ($\pm10\%$), and hue ($\pm0.08$) jitter, every parameter drawn independently per image and per epoch. Horizontal and vertical flipping are deliberately excluded: BdSL signs are articulated with respect to a dominant hand and a fixed palm orientation, so a mirrored image is not a valid example of the same class and would inject label noise rather than useful invariance. The photometric jitter is kept mild because hue and saturation shifts interact with skin tone, a genuine source of variation in the dataset rather than a nuisance to be normalized away. Augmentation is applied on-the-fly to training data only, validation and test images being rescaled but never augmented, and its contribution is quantified in Section~\ref{sec:ablationB}.\par
\subsection{Implementation and Reproducibility}
The model is implemented in TensorFlow/Keras and trained on a single NVIDIA GPU, with a deterministic image pipeline seeded at 42 and the frozen class-stratified partition of Section~\ref{sec:dataset} shared by the proposed model and every baseline. Unless stated otherwise, reported figures come from a single fixed-seed reference run; sensitivity to initialization is characterized separately by retraining from scratch over five seeds in Section~\ref{sec:mainresults}. Trained weights, training histories, and per-run metrics are archived in Keras, CSV, and JSON form so that every number reported in Section~\ref{sec:results} can be regenerated from the released artefacts.
\section{Experimental Results}
\label{sec:results}
We describe the experimental setup, report results on RSBdSL38 and six public BdSL benchmarks, quantify deployment efficiency, ablate the architecture and the training recipe, and examine what the trained model attends to, closing with a consolidated answer to the five research questions and the remaining limitations.\par
\subsection{Experimental Setup}
\label{sec:setup}
All experiments ran in a cloud environment with NVIDIA GPU acceleration. The proposed model was implemented in TensorFlow/Keras and trained with the protocol of Section~\ref{sec:methodology}: SGD with Nesterov momentum, batch size 32, input $224\times224$, reduce-on-plateau scheduling, and early stopping. The pretrained baselines were implemented in PyTorch via \texttt{timm} and fine-tuned end-to-end from ImageNet weights \cite{deng2009imagenet} with AdamW \cite{loshchilov2019decoupled} (learning rate $5\times10^{-5}$, weight decay $10^{-3}$, classifier dropout 0.3) under the same scheduling and stopping rules. The optimizer differs by design, SGD suiting a network trained from scratch and AdamW being standard for fine-tuning; every other factor is fixed, including the frozen class-stratified 8{,}794 / 977 / 1{,}103 partition of Section~\ref{sec:dataset} (80.9\% / 9.0\% / 10.1\%), the input resolution, the batch size, and the augmentation pipeline. Unless otherwise noted a single fixed seed is used; the proposed model is additionally retrained over five seeds (Section~\ref{sec:mainresults}).\par
Performance is reported as accuracy, weighted precision, weighted recall, and weighted F1-score, computed in the standard way from the per-class counts of true positives, true negatives, false positives, and false negatives. Per-class values are aggregated by support-weighted averaging; because RSBdSL38 is close to uniform, the weighted and macro averages agree to within $0.02$ percentage points, so the choice affects no conclusion below.\par
\subsection{Results on RSBdSL38}
\label{sec:mainresults}
\begin{figure}[!t]
    \centering
    \includegraphics[width=0.9\columnwidth]{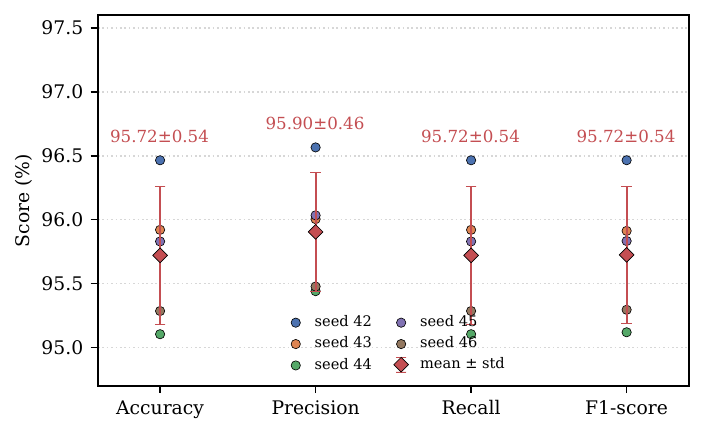}
    \caption{Five-seed stability of the proposed model on RSBdSL38, each seed retrained from scratch under the identical protocol. Coloured points are the individual seeds (42--46) and red diamonds mark the mean~$\pm$~std; the $y$-axis is zoomed to approximately 95--96.5\% so that the run-to-run spread, which is at most 1.36 percentage points, remains visible.}
    \label{fig:multiseed}
\end{figure}

Under the fixed reproducibility seed, the proposed model achieves \textbf{96.37\%} test accuracy on the 1{,}103-image RSBdSL38 test set, with weighted precision $96.50\%$, recall $96.37\%$, and F1-score $96.38\%$ (macro F1 $96.37\%$). The best checkpoint was reached at epoch~181, with training accuracy $98.56\%$ and validation accuracy $96.04\%$; the $2.52$ percentage-point train--validation gap and the close agreement between validation and test accuracy indicate that the regularization suite controls overfitting without underfitting. Retraining from scratch over five seeds (42--46) gives a mean of $95.72\% \pm 0.54\%$ accuracy, weighted precision $95.90\% \pm 0.46\%$, and F1-score $95.72\% \pm 0.54\%$; the standard deviation stays at or below $0.54$ percentage points on every metric, the full range spans only $1.36$ points ($95.10$ to $96.46\%$), and best validation accuracy is more stable still at $96.00\% \pm 0.19\%$ (Table~\ref{tab:multiseed}, Appendix~\ref{app:detailed}; Fig.~\ref{fig:multiseed}). The reference run lies $0.65$ points above the sweep mean, within $1.2$ standard deviations, so it is representative rather than exceptional; all headline comparisons use the single-run protocol applied to every model, with the sweep as the stability estimate.\par
\begin{figure}[!t]
    \centering
    \includegraphics[width=\columnwidth]{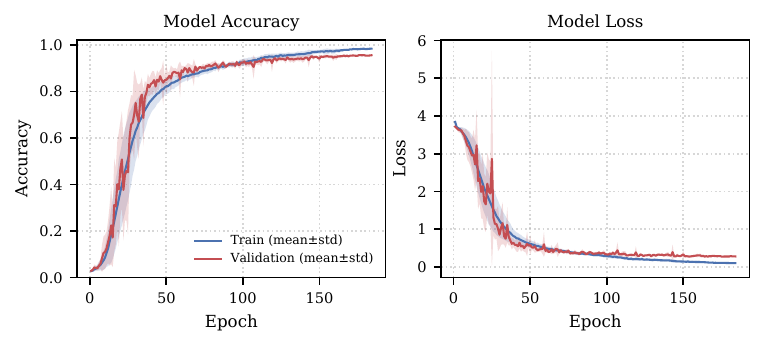}
    \caption{Training and validation accuracy (left) and loss (right) of the proposed model on RSBdSL38, averaged over five seeds (42--46). Solid lines are the mean and shaded bands denote $\pm1$~std; curves are truncated to the shortest run (184 epochs) so that every plotted epoch is a true five-seed mean. Training and validation loss track one another throughout, with no divergence indicative of overfitting.}
    \label{fig:trainhist}
\end{figure}
\begin{figure}[!t]
    \centering
    \includegraphics[width=\columnwidth]{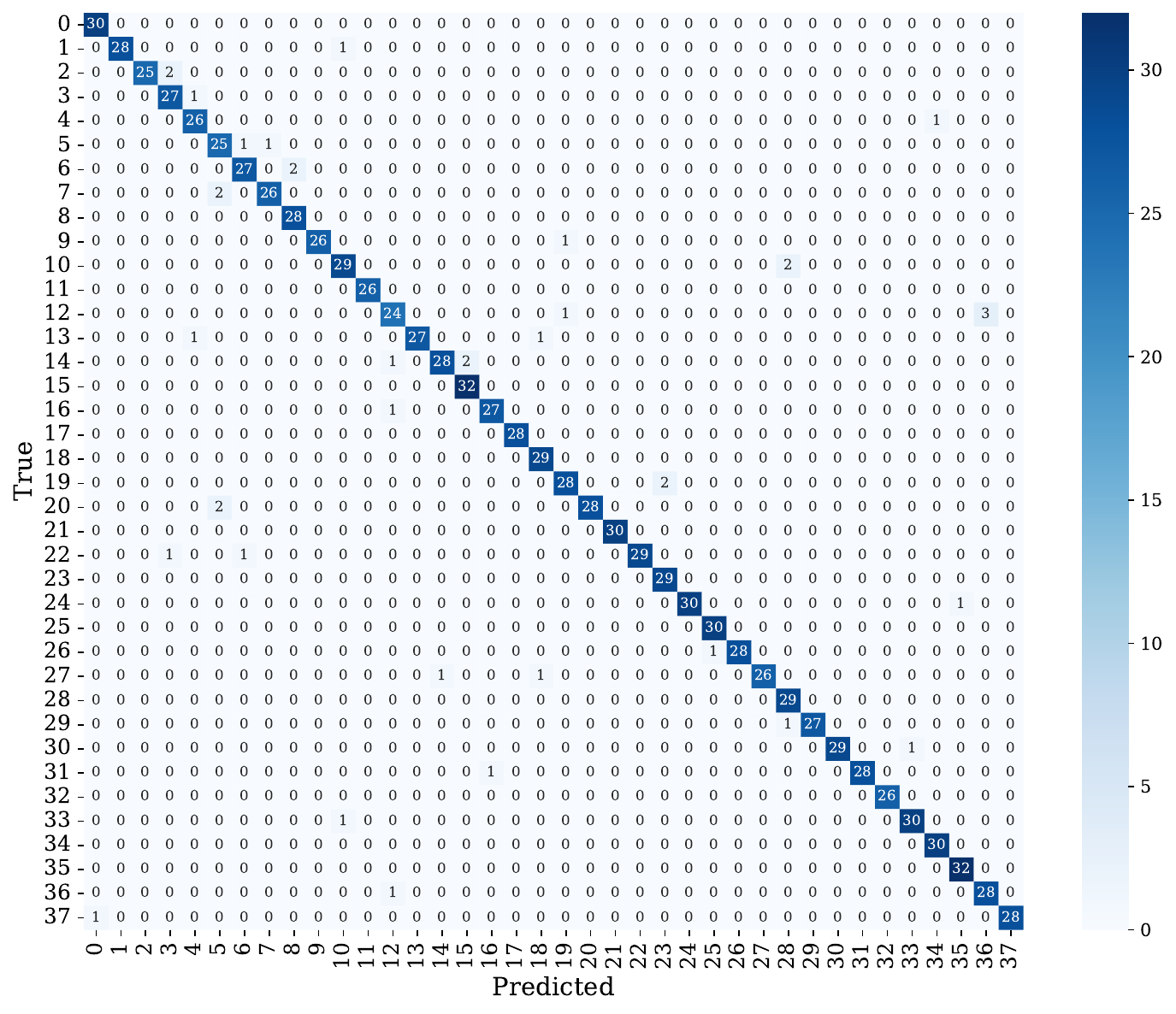}
    \caption{Confusion matrix of the proposed model on the RSBdSL38 test set (1{,}103 images, 38 classes; single-seed reference run). Forty images in total are misclassified and no off-diagonal cell exceeds three, so the residual errors are spread across many sign pairs rather than concentrated in one systematic confusion.}
    \label{fig:cm}
\end{figure}

Fig.~\ref{fig:trainhist} shows the five-seed mean training and validation curves ($\pm1$~std bands): the reduce-on-plateau schedule converges smoothly with no divergence between training and validation loss. Run length varies from 184 to 263 epochs ($206.2 \pm 34.4$) and training time from 118.9 to 177.4 minutes ($136.0 \pm 25.7$) on a single GPU, so the model reproduces in roughly two hours. Unless stated otherwise, the per-class report (Table~\ref{tab:classreport}, Appendix~\ref{app:detailed}) and the confusion matrix (Fig.~\ref{fig:cm}) correspond to the single-seed reference run. Per-class F1 ranges from 0.87 to 1.00, with four classes (11, 17, 21, 32) classified perfectly and 28 of 38 at or above 0.95; no class falls below 0.85 in precision or recall, the minima being precision $0.86$ (class~5) and recall $0.86$ (class~12). Only 40 of the 1{,}103 test images are misclassified and the errors are diffuse: the largest off-diagonal entry is three images (class~12 predicted as class~36) and every remaining pair at most two. Residual confusion concentrates among sign pairs with near-identical finger configurations, consistent with prior BdSL work \cite{hoque2020bdsl36}.\par
\subsection{Signer-Independent Evaluation}
\label{sec:signerindep}
The results so far use a class-stratified split, in which images from a given signer may appear in both partitions. This is the standard protocol in the BdSL literature, and Sections~\ref{sec:sotacomp}--\ref{sec:efficiency} adopt it so that every model is measured identically. It does, however, allow a model to exploit signer-specific cues such as hand morphology or articulation habit rather than the sign itself, inflating accuracy relative to deployment on an unseen user.\par
We therefore also evaluate under a \emph{signer-independent} protocol: of the 36 participants, 30 are assigned to training and 6 held out entirely, with validation and test images both drawn from those 6, and the network retrained from scratch under the identical protocol of Section~\ref{sec:methodology}. No test signer's images are ever seen during training, so the evaluation measures generalization to new people rather than to new images of known people. Because validation comes from the same held-out pool, checkpoint selection is exposed to those signers even though the weights are not, making the figure conservative-leaning but not fully isolated.\par
\begin{table}[!t]
  \centering
  \caption{Stratified versus signer-independent evaluation of the proposed model. The signer-independent partition holds out 6 of the 36 signers entirely from training and evaluates on their 1{,}417 images. Both models are trained from scratch under the identical protocol, so the gap isolates the effect of signer overlap.}
  \label{tab:signerindep}
  \setlength{\tabcolsep}{6pt}
  \footnotesize
  \begin{tabular}{@{}l r r r r@{}}
    \toprule
    \textbf{Protocol} & \textbf{Acc.} & \textbf{Prec.} & \textbf{Rec.} & \textbf{F1} \\
     & \textbf{(\%)} & \textbf{(\%)} & \textbf{(\%)} & \textbf{(\%)} \\
    \midrule
    Stratified (Section~\ref{sec:mainresults}) & 96.37 & 96.50 & 96.37 & 96.38 \\
    Signer-independent                         & 85.18 & 86.42 & 85.18 & 85.23 \\
    \midrule
    \textit{Gap}                               & \textit{11.19} & \textit{10.08} & \textit{11.19} & \textit{11.15} \\
    \bottomrule
  \end{tabular}
\end{table}
\begin{figure}[!t]
    \centering
    \includegraphics[width=\columnwidth]{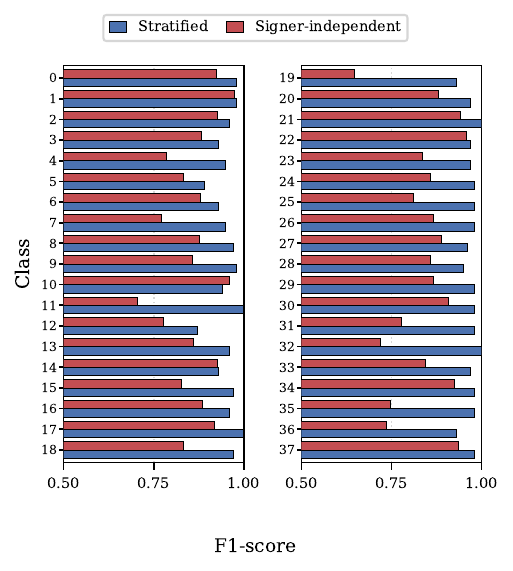}
    \caption{Per-class F1 under the stratified and the signer-independent protocols. The largest drops fall on classes whose form is most sensitive to individual hand morphology (classes 19, 11, 32, 35, and 36) rather than on the sign pairs that dominate the stratified errors, indicating that the additional errors arise from signer variation rather than from intrinsic visual ambiguity.}
    \label{fig:signerindep}
\end{figure}
Signer-independent accuracy is $85.18\%$ (weighted precision $86.42\%$, F1 $85.23\%$) over the 1{,}417 held-out test images, an $11.19$ percentage-point drop from the stratified $96.37\%$ (Table~\ref{tab:signerindep}). This is the measured cost of signer leakage: roughly a ninth of the stratified accuracy is attributable to recognizing familiar signers rather than the signs themselves. We report it openly because most published BdSL results use stratified or random splits without isolating this effect, so their headline numbers carry a comparable and usually unmeasured optimism; $85\%$ on genuinely unseen signers predicts deployed performance more faithfully than a stratified number close to the ceiling.\par
The degradation is also informative about \emph{what} fails. The train--test gap is modest ($92.19\%$ versus $85.18\%$), so the drop is not simple overfitting to the 30 training signers. The worst classes here (19, 11, 32, 36, and 35, all between $0.65$ and $0.75$ F1) differ from those dominating the stratified errors (5 and 12), so the additional errors arise from signer-specific variation in articulation rather than intrinsic ambiguity between sign pairs. Nine of the 38 classes fall below $0.80$ F1 where none do under the stratified split (Fig.~\ref{fig:signerindep}). The stratified protocol remains the primary comparison basis for consistency with the baselines and prior literature, with this figure reported alongside as the more demanding estimate.\par
\subsection{Comparison with Modern Efficient Architectures}
\label{sec:sotacomp}
Nine architectures spanning five efficient design families, MobileNetV4 \cite{qin2024mobilenetv4}, MobileViT and MobileViTv2 \cite{mehta2022mobilevit,mehta2022separable}, EfficientNetV2 \cite{tan2021efficientnetv2}, EfficientFormerV2 \cite{li2023rethinking}, and GhostNetV2 \cite{tang2022ghostnetv2}, were fine-tuned on RSBdSL38 under the identical protocol of Section~\ref{sec:setup}, with results in Table~\ref{tab:sota}.\par
\begin{table*}[!t]
  \centering
  \caption{Comparison with modern efficient architectures fine-tuned on RSBdSL38 under an identical protocol (identical splits, $224\times224$ inputs, batch size 32, identical augmentation). All pretrained models are initialized from ImageNet weights, whereas the proposed model is trained from scratch. Parameter counts are totals as reported by the respective frameworks; 292{,}262 of the proposed model's 298{,}470 parameters are trainable. Rows are ordered by accuracy. The proposed model follows the same single-run protocol as the baselines; its five-seed mean~$\pm$~std is reported in Table~\ref{tab:multiseed} and its training time is the five-seed mean.}
  \label{tab:sota}
  \setlength{\tabcolsep}{5pt}
  \footnotesize
  \begin{tabular}{@{}l r r r r r r@{}}
    \toprule
    \textbf{Model} & \textbf{Params} & \textbf{Accuracy} & \textbf{Precision} & \textbf{Recall} & \textbf{F1} & \textbf{Train time} \\
     & & \textbf{(\%)} & \textbf{(\%)} & \textbf{(\%)} & \textbf{(\%)} & \textbf{(min)} \\
    \midrule
    MobileNetV4-Hybrid-M \cite{qin2024mobilenetv4}   &  9,842,326 & 97.45 & 97.54 & 97.45 & 97.46 &  60.9 \\
    EfficientNetV2-S \cite{tan2021efficientnetv2}    & 20,226,166 & 97.15 & 97.23 & 97.15 & 97.15 &  81.8 \\
    EfficientFormerV2-S0 \cite{li2023rethinking}     &  3,259,708 & 97.09 & 97.19 & 97.09 & 97.08 &  76.7 \\
    MobileNetV4-Conv-S \cite{qin2024mobilenetv4}     &  2,541,702 & 96.97 & 97.08 & 96.97 & 96.98 &  61.2 \\
    EfficientFormerV2-S1 \cite{li2023rethinking}     &  5,752,660 & 96.91 & 96.97 & 96.91 & 96.90 &  93.5 \\
    MobileViT-S \cite{mehta2022mobilevit}            &  4,961,990 & 96.54 & 96.64 & 96.54 & 96.53 & 133.7 \\
    EfficientNetV2-B0 \cite{tan2021efficientnetv2}   &  5,907,382 & 96.48 & 96.64 & 96.48 & 96.49 &  45.0 \\
    MobileViTv2-1.0 \cite{mehta2022separable}        &  4,408,335 & 96.18 & 96.28 & 96.18 & 96.17 & 153.8 \\
    GhostNetV2-1.0 \cite{tang2022ghostnetv2}         &  4,924,586 & 95.33 & 95.44 & 95.33 & 95.32 &  66.6 \\
    \midrule
    \textbf{Proposed (from scratch)}                 & \textbf{298,470} & \textbf{96.37} & \textbf{96.50} & \textbf{96.37} & \textbf{96.38} & \textbf{136.0} \\
    \bottomrule
  \end{tabular}
\end{table*}
Three observations stand out. First, the accuracy band across all ten models is narrow (95.33 to 97.45\%), so RSBdSL38 is challenging enough that even 20M-parameter ImageNet-pretrained backbones cannot saturate it. Second, the proposed model, trained \emph{from scratch} without external data, reaches 96.37\%: above GhostNetV2-1.0 (95.33\%) and MobileViTv2-1.0 (96.18\%), within 0.11 percentage points of EfficientNetV2-B0 and 0.17 of MobileViT-S, and within 1.08 of the best model, MobileNetV4-Hybrid-M, at 33$\times$ fewer parameters. Third, in accuracy per parameter it is an order of magnitude more efficient than every baseline, the deciding property where memory, energy, and thermal budgets are tight. Against SA-CNN, the only lightweight custom BdSL architecture in the literature (671,942 parameters, 93.47\% on a comparable 38-class task) \cite{11013937}, it is 2.25$\times$ smaller and 2.90 percentage points more accurate.\par
\subsection{Cross-Dataset Generalization}
\label{sec:crossdataset}
\begin{figure}[!t]
    \centering
    \includegraphics[width=\columnwidth]{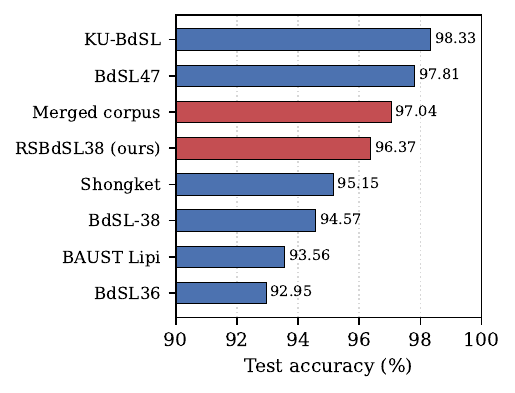}
    \caption{Test accuracy of the identical architecture retrained from scratch on six public BdSL benchmarks, on RSBdSL38, and on the merged four-dataset corpus. Accuracy exceeds 92.9\% everywhere without any per-dataset tuning, and is highest on the small, controlled corpora and lowest on those with uncontrolled backgrounds or low-resolution images.}
    \label{fig:crossdata}
\end{figure}

A central claim of this work is that the architecture is not tuned to the idiosyncrasies of its own dataset. The identical architecture, with no structural change and the same training protocol, was therefore retrained from scratch on six public BdSL benchmarks and on a merged corpus combining RSBdSL38 with BdSL-38, BdSL47, and KU-BdSL (classes aligned by a manual mapping of shared signs), as summarized in Table~\ref{tab:crossdataset} and Fig.~\ref{fig:crossdata}.\par
\begin{table}[!t]
  \centering
  \caption{Cross-dataset evaluation: the identical architecture, with no structural change and no per-dataset hyperparameter tuning, retrained from scratch on each corpus. Public benchmarks are ordered by accuracy.}
  \label{tab:crossdataset}
  \setlength{\tabcolsep}{6pt}
  \footnotesize
  \begin{tabular}{@{}l r@{}}
    \toprule
    \textbf{Dataset} & \textbf{Accuracy (\%)} \\
    \midrule
    KU-BdSL \cite{jim2023ku}               & 98.33 \\
    BdSL47 \cite{rayeed2023bdsl47}         & 97.81 \\
    Shongket \cite{hasan2021shongket}      & 95.15 \\
    BdSL-38 \cite{kabir2025combining}      & 94.57 \\
    BAUST Lipi \cite{hadiuzzaman2024baust} & 93.56 \\
    BdSL36 \cite{hoque2020bdsl36}          & 92.95 \\
    \midrule
    \textbf{RSBdSL38 (proposed dataset)}$^{\dagger}$ & \textbf{96.37} \\
    \textbf{Merged corpus (4 datasets)}              & \textbf{97.04} \\
    \bottomrule
  \end{tabular}
  \\[3pt]
  {\footnotesize $^{\dagger}$Single-seed run, consistent with the other single-run rows; the five-seed mean~$\pm$~std for RSBdSL38 is reported in Table~\ref{tab:multiseed}.}
\end{table}
The model exceeds 92.9\% on every benchmark without per-dataset tuning. On BdSL-38 it attains 94.57\% as a single 0.30M-parameter network, against 96.62\% from a five-model ensemble of heavyweight pretrained CNNs \cite{kabir2025combining} and 93.47\% from the 0.67M-parameter SA-CNN \cite{11013937}. On the merged corpus it reaches 97.04\% accuracy (weighted precision 97.06\%, F1 97.04\%), higher than on any constituent set of comparable difficulty, so the architecture benefits from rather than being confused by increased signer and environment diversity. Accuracy also tracks dataset difficulty inversely (Section~\ref{sec:litreview}): highest on the small, controlled KU-BdSL and BdSL47 sets, lowest on the uncontrolled backgrounds of BdSL36 and the low-resolution BAUST Lipi images.\par
\subsection{Zero-Shot Cross-Dataset Transfer}
\label{sec:zeroshot}
Section~\ref{sec:crossdataset} retrains the architecture on each target dataset and so tests whether the \emph{design} transfers. A stricter question is whether the \emph{trained model} transfers with no adaptation at all, which is what a deployed model faces on data from a new source and is rarely reported in the BdSL literature. We evaluate the RSBdSL38-trained reference model directly on the full public distribution of BdSL-38 \cite{kabir2025combining} (12{,}581 images, identical 38-class label scheme, verified one-to-one) without any fine-tuning.\par
The model attains $76.25\%$ zero-shot accuracy (macro F1 $76.00\%$, weighted F1 $76.03\%$). The contrast with the $94.57\%$ obtained when the same architecture is retrained on BdSL-38 (Table~\ref{tab:crossdataset}) isolates the domain shift: the $18.32$ percentage-point difference is the portion of BdSL-38 accuracy that depends on adapting to its specific signers, cameras, and capture conditions rather than on the cross-corpus invariant structure of the signs. That a model which has never seen a BdSL-38 image classifies three quarters of it correctly indicates substantially signer- and source-agnostic representations rather than memorized dataset artifacts.\par
\begin{figure}[!t]
    \centering
    \includegraphics[width=\columnwidth]{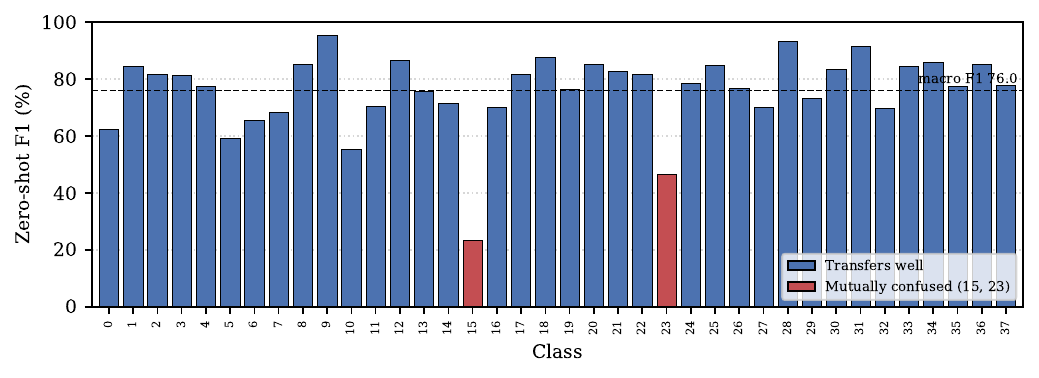}
    \caption{Per-class F1 for zero-shot transfer of the RSBdSL38-trained model to BdSL-38, with no fine-tuning. Most classes transfer well, with 34 of 38 above $0.60$~F1; the two low outliers (classes 15 and 23) are mutually confused rather than failing independently.}
    \label{fig:zeroshot}
\end{figure}
The per-class breakdown (Fig.~\ref{fig:zeroshot}) shows broadly even transfer, 34 of 38 classes above $0.60$~F1 and a majority above $0.75$, but two degrade sharply: class~15 ($0.23$~F1) and class~23 ($0.47$). These are not independent failures: of the 340 class-15 test images, 191 are predicted as class~23, and 93 class-23 images as class~15. Because the label mapping was verified beforehand, this is a genuine visual near-degeneracy in how BdSL-38 renders the two articulations rather than an annotation mismatch, and the retrained result confirms it is recoverable with a little target-domain fine-tuning. We report the zero-shot figure as a lower bound on deployed performance; it is not comparable to the retrained baselines of Section~\ref{sec:sotacomp} and is used in no accuracy comparison against them.\par
\subsection{Deployment Efficiency}
\label{sec:efficiency}
\begin{figure}[!t]
    \centering
    \includegraphics[width=0.8\columnwidth]{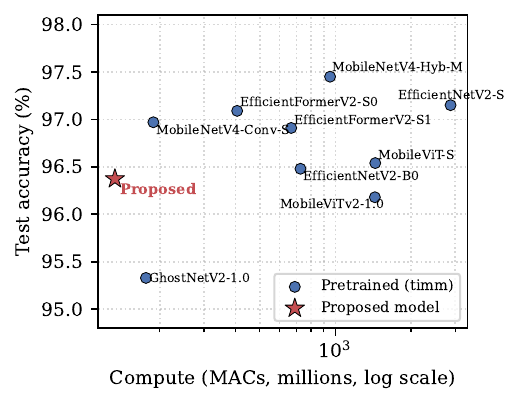}
    \caption{Accuracy versus compute cost (MACs, logarithmic scale) on RSBdSL38. The proposed model (star) has the lowest compute cost of the ten models compared, and no model within 1.1 percentage points of its accuracy costs fewer multiply--accumulate operations.}
    \label{fig:efficiency}
\end{figure}

Parameter count alone does not determine deployability; compute cost, latency, and on-disk size matter as much on resource-constrained hardware. This section reports all three for the proposed model against the nine pretrained baselines of Section~\ref{sec:sotacomp} on identical $224\times224$ inputs, summarized in Table~\ref{tab:efficiency} and plotted against compute cost in Fig.~\ref{fig:efficiency}.\par
\begin{table*}[!t]
  \centering
  \caption{Deployment efficiency of the proposed model against the pretrained baselines. Multiply--accumulate operations (MACs) and FP32 size are measured at $224\times224$. The final row gives the baseline-to-proposed ratio, from the least to the most demanding baseline on each axis; larger is better for the proposed model. Quantized sizes and measured on-device latency are reported for the proposed model in the text and in Table~\ref{tab:ondevice}.}
  \label{tab:efficiency}
  \setlength{\tabcolsep}{6pt}
  \footnotesize
  \begin{tabular}{@{}l r r r r@{}}
    \toprule
    \textbf{Model} & \textbf{Params} & \textbf{MACs (M)} & \textbf{FP32 size (MB)} & \textbf{Accuracy (\%)} \\
    \midrule
    MobileNetV4-Hybrid-M \cite{qin2024mobilenetv4}   &  9,842,326 &  952.7 & 37.5 & 97.45 \\
    EfficientNetV2-S \cite{tan2021efficientnetv2}    & 20,226,166 & 2873.0 & 77.2 & 97.15 \\
    EfficientFormerV2-S0 \cite{li2023rethinking}     &  3,259,708 &  406.7 & 12.4 & 97.09 \\
    MobileNetV4-Conv-S \cite{qin2024mobilenetv4}     &  2,541,702 &  188.8 &  9.7 & 96.97 \\
    EfficientFormerV2-S1 \cite{li2023rethinking}     &  5,752,660 &  667.4 & 21.9 & 96.91 \\
    MobileViT-S \cite{mehta2022mobilevit}            &  4,961,990 & 1441.3 & 18.9 & 96.54 \\
    EfficientNetV2-B0 \cite{tan2021efficientnetv2}   &  5,907,382 &  726.4 & 22.5 & 96.48 \\
    MobileViTv2-1.0 \cite{mehta2022separable}        &  4,408,335 & 1436.3 & 16.8 & 96.18 \\
    GhostNetV2-1.0 \cite{tang2022ghostnetv2}         &  4,924,586 &  176.3 & 18.8 & 95.33 \\
    \midrule
    \textbf{Proposed (from scratch)} & \textbf{298,470} & \textbf{132.7} & \textbf{1.14} & \textbf{96.37} \\
    \textit{\quad ratio (min--max)}  & \textit{8.5--68$\times$} & \textit{1.3--21.7$\times$} & \textit{8.5--68$\times$} & --- \\
    \bottomrule
  \end{tabular}
\end{table*}
\textbf{Compute.} At $132.7$~M MACs the proposed model is the cheapest in the comparison, needing $1.3\times$ fewer MACs than the next-lightest baseline (GhostNetV2-1.0) and $21.7\times$ fewer than the heaviest (EfficientNetV2-S). The two MobileViT transformer baselines cost roughly $11\times$ its compute; of these, only MobileViT-S is more accurate, and by just $0.17$ percentage points ($96.54\%$ versus $96.37\%$), while MobileViTv2-1.0 is in fact $0.19$ points behind it. In either case the compute penalty is far larger than the accuracy difference.\par
\textbf{On-disk size.} As a single-precision Keras model the network occupies $1.14$~MB, $8.5$ to $68\times$ smaller than the baselines. Post-training quantization compresses it further with negligible loss: dynamic-range TensorFlow Lite (TFLite) to $0.43$~MB and full-integer INT8 to $0.46$--$0.48$~MB, the latter actually \emph{improving} test accuracy fractionally to $96.46\%$. At under half a megabyte it fits the flash budget of low-cost microcontrollers, which no pretrained baseline approaches.\par
\textbf{Latency.} Measured single-image inference latency is $10.2 \pm 0.8$~ms on GPU (batch size 1) and $11.8$~ms on desktop CPU through the XNNPACK delegate. Because latency depends on the measurement platform, we report it only for the proposed model and use MACs as the hardware-independent compute proxy in Fig.~\ref{fig:efficiency}.\par
\textbf{On-device deployment.} We deployed the $0.48$~MB INT8 TFLite model on a commodity Android smartphone (Snapdragon~7+~Gen~3, 12~GB RAM) with the XNNPACK delegate and four threads. Over 249 timed runs the mean latency is $3.98 \pm 0.09$~ms (median $3.96$, 95th percentile $4.16$), approximately $251$~frames per second, with peak memory $15.5$~MB, initialization $13.8$~ms, and $241$ of $315$ operators hardware-delegated (Table~\ref{tab:ondevice}, Appendix~\ref{app:ablation}). This exceeds the desktop-CPU figure because integer quantization matches the vector units of the mobile system-on-chip (SoC), establishing real-time recognition within a mobile memory budget without a GPU.\par
Together these measurements substantiate the deployability claim in the paper's title: the model is not merely small in parameter count but cheap to compute, sub-megabyte on disk, and real-time on a commodity smartphone, within $1.1$ percentage points of backbones one to two orders of magnitude larger on every efficiency axis.\par
\subsection{Ablation Study A: Stage-Wise Depth Ablation}
\label{sec:ablationA}
The first ablation asks how much each of the four stages contributes. We evaluate all 14 non-empty stage-removal configurations (four single, six pairwise, three triple, and all four) and retrain each from scratch under the full protocol of Section~\ref{sec:methodology} on the identical split. All comparisons use one reference point, the full model D0, which reached $96.55\%$ with 298{,}470 parameters in 192 epochs; this is an independent run of the same configuration as the reference model of Section~\ref{sec:mainresults} ($96.37\%$), the $0.18$-point difference lying well inside the $\pm0.54$ seed variability of Table~\ref{tab:multiseed}, and fixing one baseline lets this study and Section~\ref{sec:ablationB} share an operating point. Multi-stage runs retain the fixed $2\times2$ downsampling transitions of Stages~1 and~2 so that downstream spatial resolution stays comparable; single-stage runs follow their recorded stage-skip definitions, including removal of the stage-local pooling when Stage~1 or Stage~2 is omitted. Fig.~\ref{fig:stageablation} summarizes the results, with per-variant metrics in Table~\ref{tab:stageablation} of Appendix~\ref{app:ablation}.\par
\begin{figure}[!t]
    \centering
    \includegraphics[width=\columnwidth]{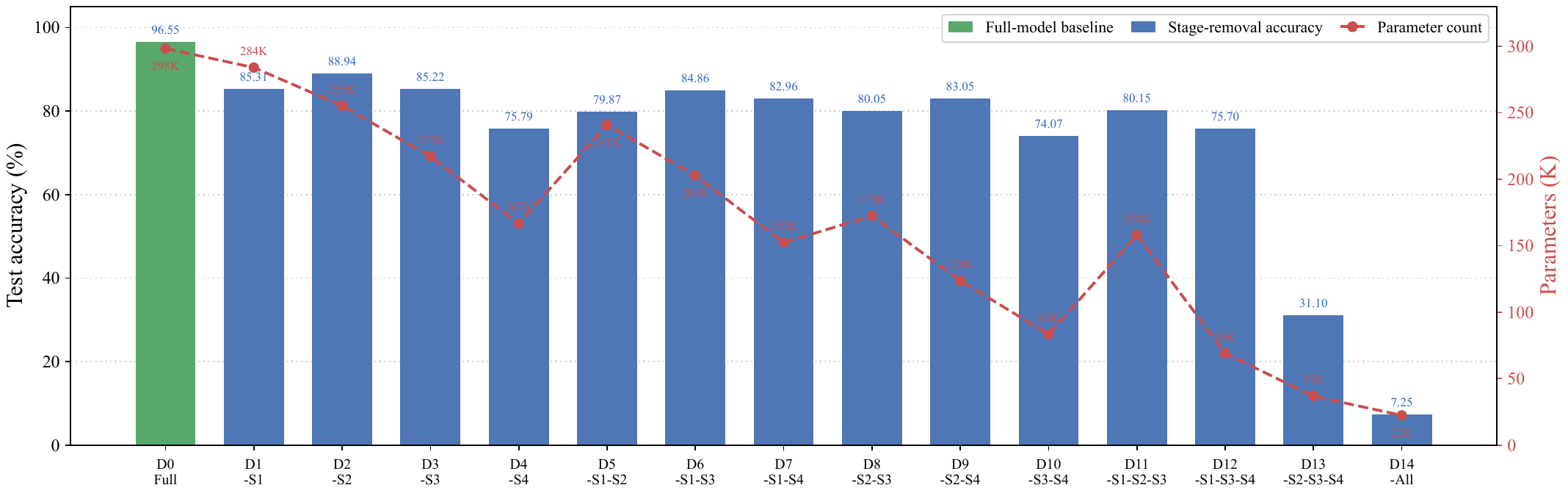}
    \caption{Stage-wise depth ablation relative to the common D0 baseline. Bars show test accuracy for each of the 14 stage-removal configurations and the dashed line shows the corresponding parameter count. Every configuration degrades substantially, and the ordering of the bars does not follow the ordering of the parameter counts, showing that which stages remain matters more than how many parameters remain.}
    \label{fig:stageablation}
\end{figure}
Every single-stage removal produces a material loss. Stage~2 is the least damaging omission: D2 reaches $88.94\%$, $7.61$ percentage points below D0, for a $14.4\%$ parameter saving. Removing Stage~1 or Stage~3 costs $11.24$ and $11.33$ points respectively, and Stage~4 is the most important: D4 falls to $75.79\%$, a $20.76$ percentage-point deficit. The late 128-filter stage therefore contributes information the earlier stages cannot recover on their own.\par
The pairwise results reveal interactions hidden by the single-stage runs. D6 (without S1 and S3) is the strongest pairwise removal at $84.86\%$, whereas removing both late stages (D10) reduces accuracy to $74.07\%$. Parameter count does not explain this ordering: D9 retains only 123{,}686 parameters ($58.6\%$ fewer than D0) yet reaches $83.05\%$, while D5 has 241{,}030 parameters but reaches only $79.87\%$. Which stages remain therefore matters more than raw model size.\par
The triple and all-stage removals clarify the hierarchy: retaining only Stage~4 (D11) gives $80.15\%$, only Stage~2 (D12) $75.70\%$, and only Stage~1 (D13) $31.10\%$, while D14, which keeps only the stem, the fixed transitions, the pooling fusion, and the classifier head, reaches $7.25\%$. Shallow features alone are therefore insufficient, and the deeper stages, especially Stage~4, carry most of the class-discriminative representation; Fig.~\ref{fig:ablationcurves} in Appendix~\ref{app:ablation} agrees, the severely reduced variants converging at substantially higher loss.\par
For deployment the study exposes a genuine trade-off rather than a nearly free reduction: D2 is the highest-accuracy reduced configuration but still $7.61$ percentage points down, and D9 retains $83.05\%$ at $58.6\%$ fewer parameters, so no stage-removal variant remains close to D0. We therefore keep the full four-stage network as the reference architecture and rely on post-training quantization, rather than architectural trimming, to reach the sub-megabyte deployment target of Section~\ref{sec:efficiency}.\par
\subsection{Ablation Study B: Component Controls}
\label{sec:ablationB}
\begin{figure}[!t]
    \centering
    \includegraphics[width=\columnwidth]{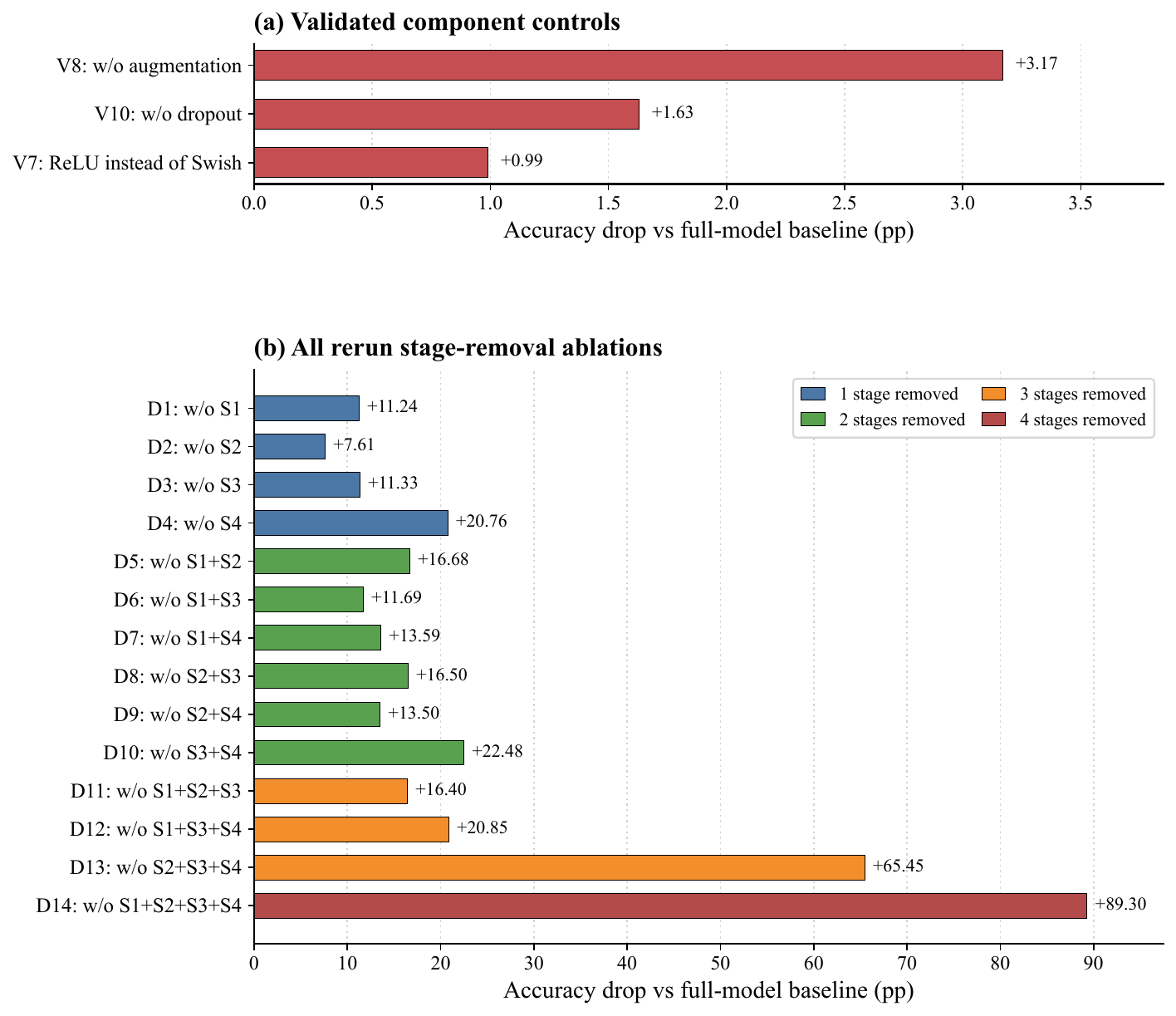}
    \caption{Consolidated accuracy degradation relative to the common full-model baseline ($96.55\%$). Panel (a) contains the three parameter-matched component controls and panel (b) contains all 14 stage-removal configurations (D1--D14). The two panels use different horizontal ranges so that the smaller component effects remain legible, so cross-panel comparisons should be read from the printed bar-end values rather than from bar lengths. The baseline has zero degradation and is therefore not drawn as a bar.}
    \label{fig:compablation}
\end{figure}

The second ablation isolates the training recipe and the activation function from model capacity. Three controls are evaluated against the same full-model baseline, relabelled V0: no data augmentation, no dropout, and ReLU in place of Swish. All three share V0's exact 298{,}470-parameter architecture, so their differences must reflect the removed mechanism rather than capacity. Table~\ref{tab:compablation} reports the four configurations and Fig.~\ref{fig:compablation} places them alongside all 14 stage removals against the common $96.55\%$ baseline; variant identifiers follow the experiment log so that the figure, the table, and the released artefacts can be cross-referenced.\par
\begin{table}[!t]
  \centering
  \caption{Parameter-matched component controls. All variants share the 298{,}470-parameter architecture of V0, so the differences isolate the removed mechanism rather than a change in capacity. Metrics are percentages and $\Delta$ is the accuracy degradation relative to V0.}
  \label{tab:compablation}
  \setlength{\tabcolsep}{4pt}
  \scriptsize
  \begin{tabular}{@{}l r r r r r r@{}}
    \toprule
    \textbf{Variant} & \textbf{Acc.} & \textbf{Prec.} & \textbf{Rec.} & \textbf{F1} & \textbf{Ep.} & \textbf{$\Delta$} \\
     & \textbf{(\%)} & \textbf{(\%)} & \textbf{(\%)} & \textbf{(\%)} & & \textbf{(pp)} \\
    \midrule
    \textbf{V0: Full model}   & \textbf{96.55} & \textbf{96.65} & \textbf{96.55} & \textbf{96.55} & \textbf{192} & \textbf{0.00} \\
    V8: w/o augmentation      & 93.38 & 93.51 & 93.38 & 93.37 & 234 & $+3.17$ \\
    V10: w/o dropout          & 94.92 & 95.10 & 94.92 & 94.90 & 111 & $+1.63$ \\
    V7: ReLU instead of Swish & 95.56 & 95.71 & 95.56 & 95.58 & 198 & $+0.99$ \\
    \bottomrule
  \end{tabular}
\end{table}
Data augmentation has the largest effect: removing it drops accuracy from $96.55\%$ to $93.38\%$ ($-3.17$ percentage points) and runs 42 epochs longer before early stopping, consistent with a model still fitting an unaugmented set it has begun to memorize. Removing dropout lowers accuracy to $94.92\%$ ($-1.63$) and cuts the run to 111 epochs, the shortest of the four and the signature of premature convergence onto a sharper minimum. Replacing Swish with ReLU gives the smallest but still consistent reduction, to $95.56\%$ ($-0.99$), supporting the smooth activation of the full model \cite{ramachandran2017swish}. Because the parameter count is identical, none of these gaps reflects model size.\par
Fig.~\ref{fig:compablation} clarifies the relative scale: the component controls span $0.99$ to $3.17$ percentage points, the 14 stage removals $7.61$ to $89.30$, and even D2, the least damaging stage removal, costs $2.4\times$ what removing augmentation does. Preserving the full stage hierarchy therefore matters far more, though augmentation, dropout, and Swish each add measurable gains, so neither the four-stage extractor nor the regularized recipe can be dispensed with to reproduce the full-model operating point.\par
\subsection{Explainability Analysis}
\label{sec:explainability}
The ablation studies establish which parts of the network matter for accuracy, but not \emph{what} it attends to. Because the deployment setting involves uncontrolled backgrounds, we must confirm that predictions follow the signing hand rather than incidental scene context, a failure mode accuracy cannot detect (RQ5). We combine qualitative attribution maps with quantitative faithfulness measures and a sanity check.\par
\textbf{(i) Method.} We apply Grad-CAM \cite{selvaraju2017grad} and its generalization Grad-CAM++ \cite{chattopadhay2018grad} to the output of the Stage-4 hand-feature block, the deepest $14 \times 14 \times 128$ representation that still retains spatial structure. Let $A^k$ denote the $k$-th channel of that feature map and $y^c$ the pre-softmax score of class $c$. Grad-CAM weights each channel by its spatially averaged gradient,
\begin{equation}
\alpha^c_k = \frac{1}{H'W'} \sum_{i} \sum_{j} \frac{\partial y^c}{\partial A^k_{ij}},
\label{eq:gradcam-alpha}
\end{equation}
and forms the localization map as the rectified weighted combination
\begin{equation}
L^c = \mathrm{ReLU}\!\left( \sum_{k} \alpha^c_k \, A^k \right),
\label{eq:gradcam-map}
\end{equation}
which is min--max normalized and bilinearly upsampled to the $224 \times 224$ input grid. Grad-CAM++ replaces the uniform spatial average in Eq.~\eqref{eq:gradcam-alpha} with a positive, pixel-wise weighting derived from higher-order derivatives of $y^c$, which improves localization when several disjoint regions support the same class.\par
\begin{figure}[!t]
    \centering
    \includegraphics[width=\columnwidth]{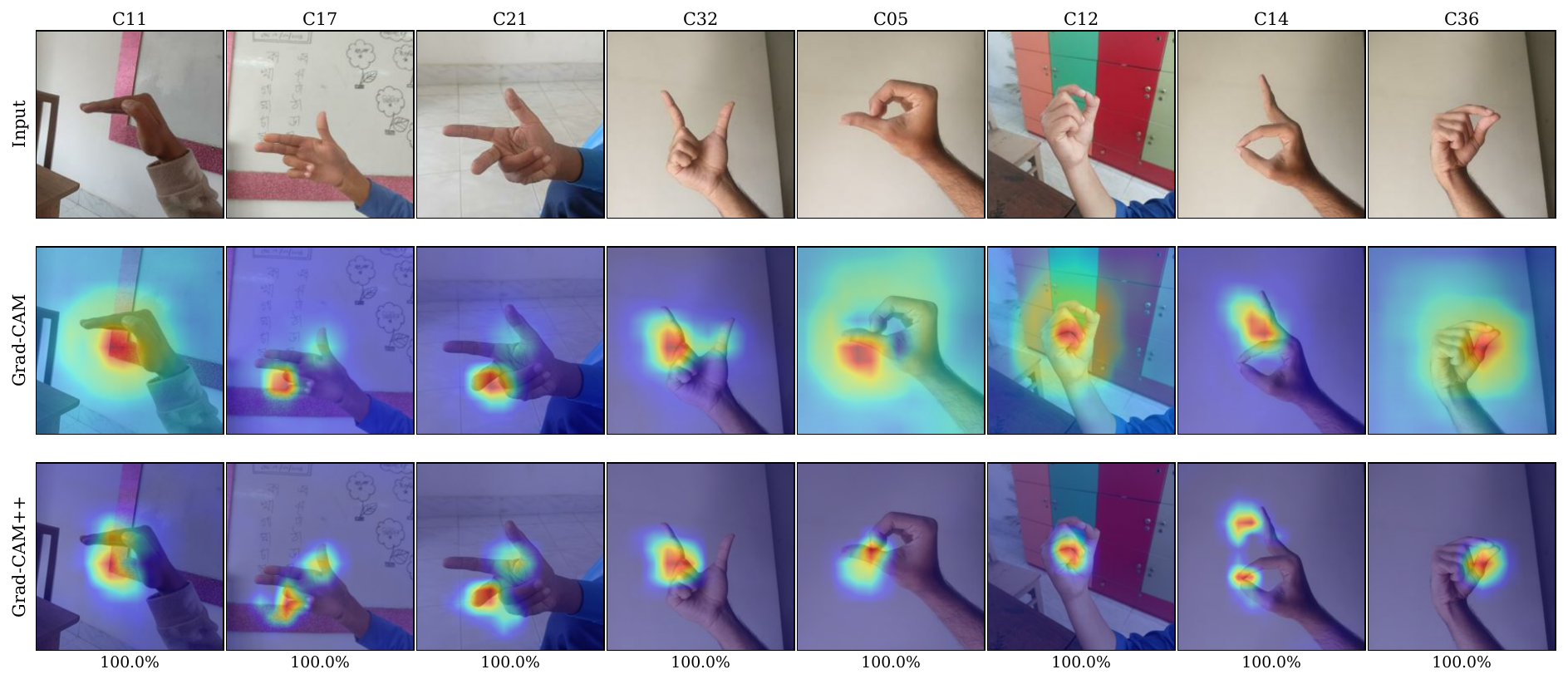}
    \caption{Grad-CAM (middle row) and Grad-CAM++ (bottom row) localization maps for eight RSBdSL38 test images, the four highest-F1 and the four lowest-F1 classes, with the predicted-class probability printed beneath each map. The attribution peak falls on the configured fingers in every case, including the four samples captured against high-contrast structured backgrounds.}
    \label{fig:gradcam-grid}
\end{figure}

\textbf{(ii) Qualitative attribution.} Fig.~\ref{fig:gradcam-grid} shows maps for the four highest-F1 classes in Table~\ref{tab:classreport} (11, 17, 21, 32, all at 1.00) and the four lowest (5, 12, 14, 36), each predicted with essentially full confidence. The attribution peak falls on the signing hand in all eight cases, specifically on the configured fingers rather than the wrist or forearm, which are shared across classes and carry no discriminative signal.\par
The behaviour under adverse capture conditions is more informative, because four of the eight samples were photographed against strongly structured backgrounds: saturated red, green, and pink lockers (class~12), a whiteboard covered in handwritten Bangla characters (class~17), a tiled floor with regular grout lines (class~21), and a wall crossed by a magenta stripe with furniture in frame (class~11). In each case the peak stays on the hand and the competing structure draws no comparable response, even though the background occupies most of the frame and, for the lockers and whiteboard, is higher in contrast than the hand. This is direct evidence that the spatial attention module suppresses background as intended in Section~\ref{sec:methodology}, and helps explain the transfer to uncontrolled capture conditions without per-dataset tuning (Section~\ref{sec:crossdataset}).\par
\begin{figure}[!t]
    \centering
    \includegraphics[width=\columnwidth]{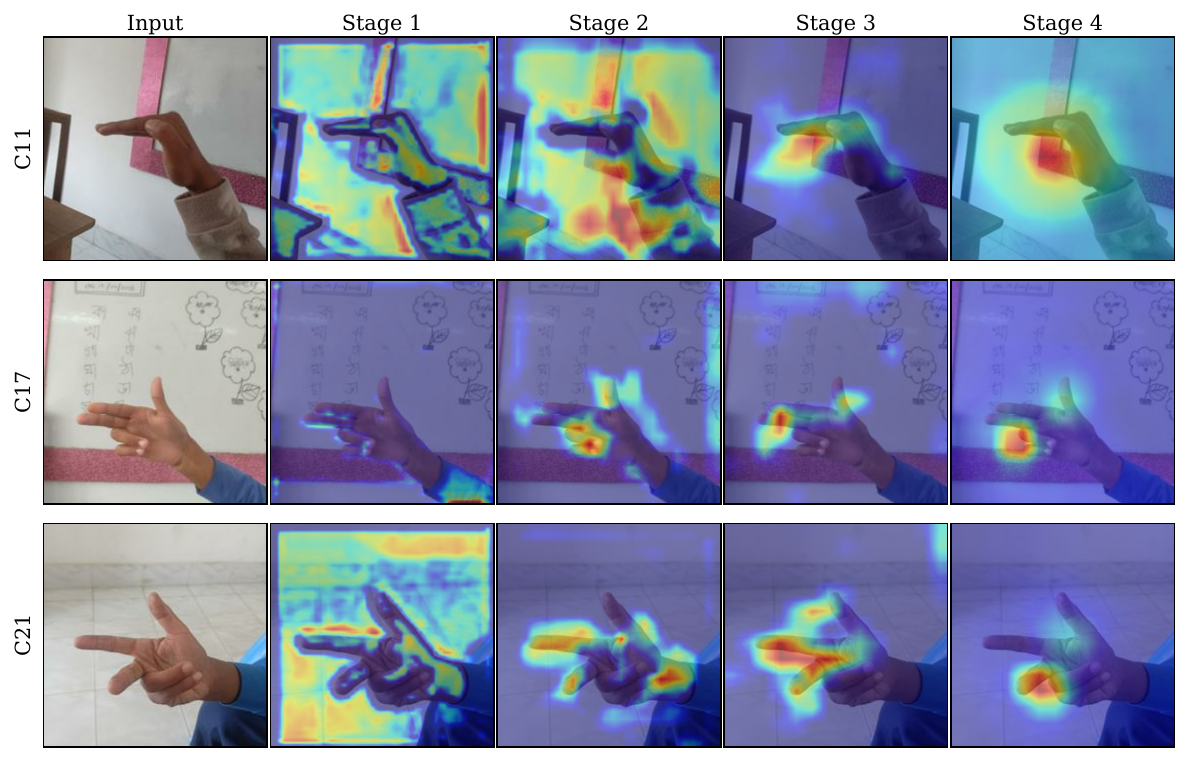}
    \caption{Stage-wise evolution of the Grad-CAM map for three classes. Attribution moves from a generic edge response at Stage~1, through partial figure--ground separation at Stage~2 and sharp per-finger peaks at Stage~3, to a single consolidated hand region at Stage~4.}
    \label{fig:gradcam-stage}
\end{figure}

\textbf{(iii) Where the attribution refines.} Fig.~\ref{fig:gradcam-stage} traces the map through the four stages for classes 11, 17, and 21. Stage~1 behaves essentially as an edge detector, responding along every contour with no preference for the hand; Stage~2 begins to separate hand from background but retains substantial off-hand response; Stage~3 produces sharp, multi-modal peaks on individual extended fingers; Stage~4 consolidates these into a single region covering the discriminative finger--palm configuration. Background rejection is therefore established between Stages~2 and~3, consistent with the depth ablation of Section~\ref{sec:ablationA}, where removing Stage~2 or Stage~3 costs $7.61$ and $11.33$ percentage points even though both leave the deepest stage intact, and with the design intent of Section~\ref{sec:methodology}.\par
\begin{figure}[!t]
    \centering
    \includegraphics[width=\columnwidth]{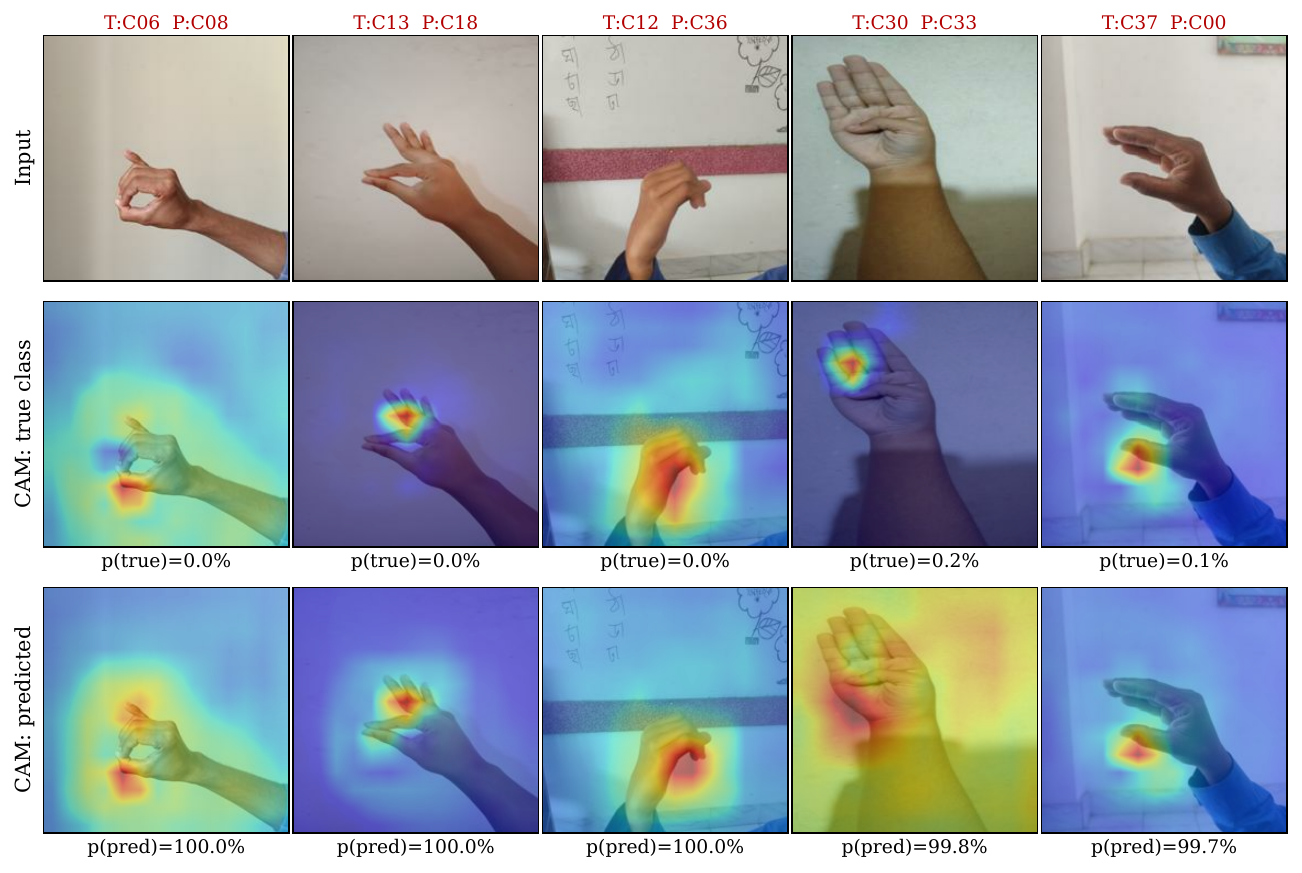}
    \caption{Failure analysis of five misclassified test images. For each image (top row), the Grad-CAM map for the ground-truth class (middle row) and for the predicted class (bottom row) are shown with the corresponding class probabilities. Both maps fall on the same hand region in every case, indicating a discrimination failure rather than a localization failure.}
    \label{fig:gradcam-error}
\end{figure}

\textbf{(iv) Failure analysis.} Fig.~\ref{fig:gradcam-error} renders, for five misclassified test images, the map for the ground-truth class alongside that for the predicted class. In all five the two maps fall on the same hand, so the model localizes correctly and fails at discrimination rather than at attention. What separates them is extent rather than position: the ground-truth map is tight on the finger detail that distinguishes the true sign, the predicted-class map broader, covering the palm and the adjacent finger group. The third column is the dominant error mode, class~12 predicted as class~36, where both maps sit on the same fist-and-thumb configuration and the distinction is finer than the $14 \times 14$ attribution grid can resolve.\par
The confidence distribution over the 40 errors corroborates this: mean probability $0.81$ on the wrong class, 19 of the 40 above $0.9$ and 11 above $0.99$, against only $0.09$ on the true class, with just 5 errors exceeding $0.3$. The model is confidently rather than marginally wrong, so with the localization evidence the residual errors arise from genuine visual near-degeneracy between sign pairs, most prominently classes 12 and 36, rather than from background distraction. Higher input or feature resolution, not stronger attention, is the appropriate remedy; the errors persist at full four-stage capacity, so they are not a symptom of insufficient depth.\par
\textbf{(v) Faithfulness.} Visual plausibility does not establish that a map reflects the evidence the network uses. We compute the deletion and insertion measures of Petsiuk et al. \cite{petsiuk2018rise} over 200 test images, progressively removing pixels from or restoring them to a blurred baseline in order of decreasing attribution and recording the area under the resulting predicted-probability curve (AUC); a faithful map gives a low deletion AUC and a high insertion AUC. We also report the average confidence drop and increase-in-confidence rate of Chattopadhay et al. \cite{chattopadhay2018grad}, obtained by masking the input with the normalized map (Table~\ref{tab:gradcam}).\par
\begin{table}[!t]
  \centering
  \caption{Faithfulness of the attribution maps over 200 test images. Lower deletion AUC, higher insertion AUC, lower average confidence drop, and higher increase-in-confidence rate indicate a more faithful explanation. The random control ranks pixels arbitrarily and bounds the performance of an uninformative map.}
  \label{tab:gradcam}
  \setlength{\tabcolsep}{5pt}
  \scriptsize
  \begin{tabular}{@{}l c c c c@{}}
    \toprule
    \textbf{Method} & \textbf{Del.\ AUC} & \textbf{Ins.\ AUC} & \textbf{Avg.\ drop} & \textbf{Increase} \\
     & $\downarrow$ & $\uparrow$ & \textbf{(\%)} $\downarrow$ & \textbf{(\%)} $\uparrow$ \\
    \midrule
    Random (control)  & 0.211 & 0.605 & --- & --- \\
    Grad-CAM++        & 0.110 & 0.774 & 97.40 & 0.50 \\
    \textbf{Grad-CAM} & \textbf{0.081} & \textbf{0.888} & \textbf{53.71} & \textbf{7.50} \\
    \bottomrule
  \end{tabular}
\end{table}
\begin{figure}[!t]
    \centering
    \includegraphics[width=\columnwidth]{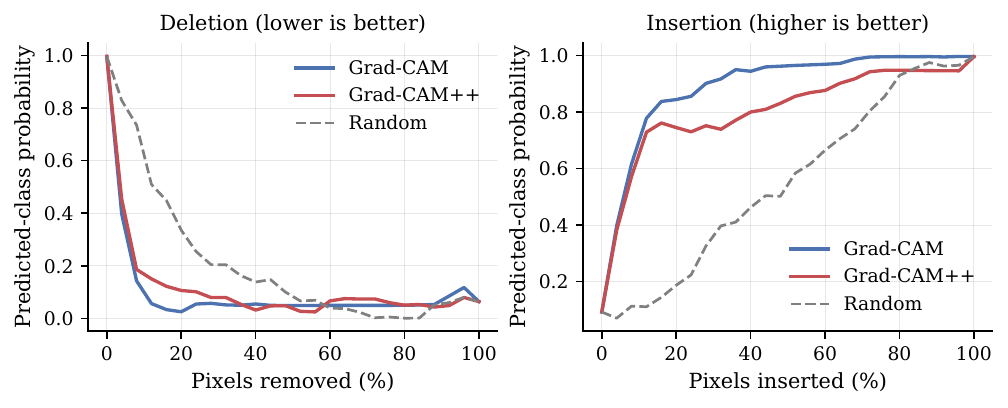}
    \caption{Deletion (left) and insertion (right) faithfulness curves for Grad-CAM, Grad-CAM++, and a random control, averaged over 200 test images. Pixels are removed from (deletion) or restored to (insertion) a blurred baseline in order of decreasing attribution. A faithful map drives the predicted-class probability down fastest, giving the smallest area under the deletion curve, and recovers it soonest, giving the largest area under the insertion curve; Grad-CAM dominates on both criteria and separates clearly from the random baseline.}
    \label{fig:deletion-insertion}
\end{figure}
Fig.~\ref{fig:deletion-insertion} plots the full curves. Both attribution methods separate clearly from the random control, confirming that the highlighted region is the evidence the network relies on rather than a plausible-looking overlay: deleting pixels in Grad-CAM order collapses the predicted-class probability roughly $2.6\times$ faster than random deletion (deletion AUC $0.081$ versus $0.211$), and restoring them recovers the prediction far sooner (insertion AUC $0.888$ versus $0.605$).\par
Grad-CAM \cite{selvaraju2017grad} is the more faithful of the two on every measure, and the gap is explained by map sparsity, visible in Fig.~\ref{fig:gradcam-grid}: the Grad-CAM++ maps are compact and tightly bounded to the fingers, the Grad-CAM maps extend over the whole hand and somewhat beyond. The sharper maps are the less faithful, because restricting the input to the Grad-CAM++ region discards contextual evidence the network in fact uses, producing near-total confidence collapse under masking ($97.40\%$ average drop). Since the compact backbone aggregates evidence over the whole hand rather than a few isolated keypoints, the broader maps describe its behaviour more accurately, and all qualitative analysis here uses Grad-CAM.\par
\begin{figure}[!t]
    \centering
    \includegraphics[width=\columnwidth]{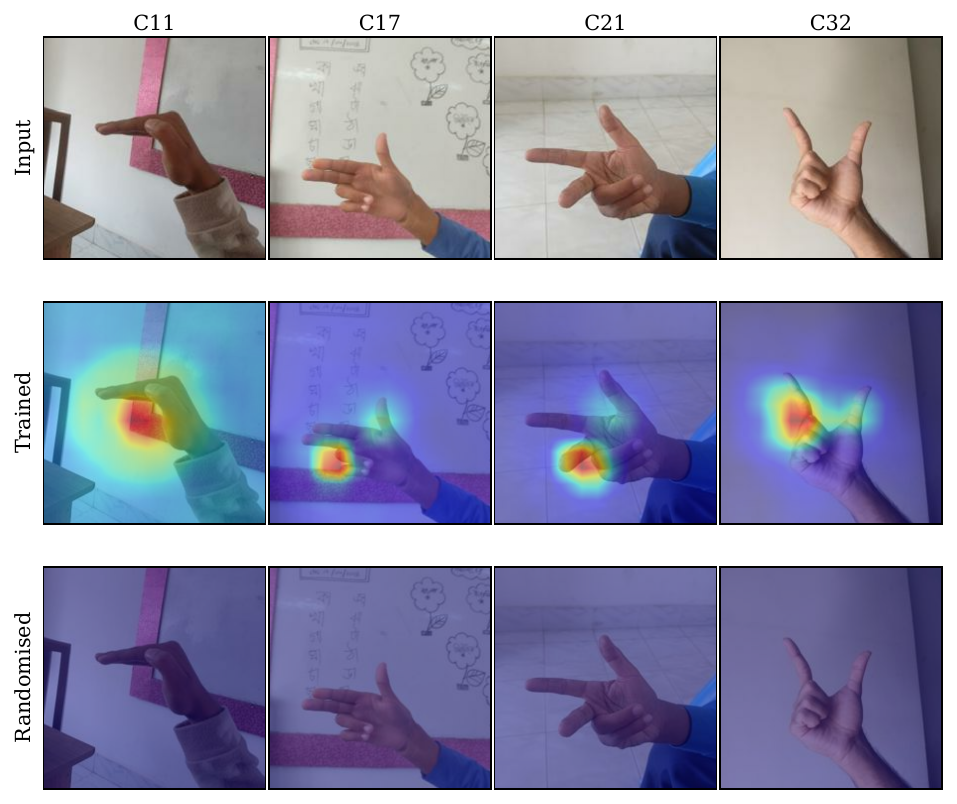}
    \caption{Sanity check by cascading weight randomization \cite{adebayo2018sanity}. The input images (top), the Grad-CAM maps of the trained model (middle), and the maps obtained after randomizing the final stage and the classifier head (bottom). The randomized maps collapse to a near-uniform field, so the localization is a property of the learned weights rather than of the input's edge content.}
    \label{fig:gradcam-sanity}
\end{figure}

\textbf{(vi) Sanity check.} Following Adebayo et al. \cite{adebayo2018sanity}, we verify that the attribution depends on the learned parameters rather than acting as an edge detector. After cascading randomization of the classifier head and the final stage, the maps collapse to a near-uniform field with no correspondence to the hand (Fig.~\ref{fig:gradcam-sanity}), unlike the sharply localized maps of the trained model. They notably do not fall back on the contour structure that dominates Stage~1, confirming that localization is a property of the learned weights rather than of the input's edge content, so the explanations satisfy a model-sensitivity criterion that many published saliency analyses fail.\par
The analysis is subject to one methodological caveat. Because Stage~4 operates on a $14 \times 14$ grid, each map cell corresponds to a $16 \times 16$ input patch, so the maps identify the discriminative region but cannot resolve individual finger boundaries; the Stage-3 maps in Fig.~\ref{fig:gradcam-stage} are correspondingly sharper.\par
\subsection{Summary of Findings}
\label{sec:discussion}
Careful architectural design, attention, multi-scale depthwise features, and aggressive parameter sharing, lets a from-scratch 0.30M-parameter model operate within striking distance of ImageNet-pretrained backbones 8.5 to 68$\times$ its size. On \textbf{RQ1}, it attains 96.37\% on RSBdSL38 (five-seed mean $95.72\% \pm 0.54\%$), ahead of two pretrained baselines and within 1.08 percentage points of the best, while being the most compute- and memory-efficient model in the suite (Section~\ref{sec:efficiency}).\par
On \textbf{RQ2}, the identical architecture exceeds 92.9\% on all six public benchmarks and reaches 97.04\% on the merged corpus, so the design is not tuned to its own dataset and in fact benefits from increased signer and environment diversity; zero-shot transfer to BdSL-38 at 76.25\% shows its representations to be largely source-agnostic.\par
For \textbf{RQ3}, the two ablation studies separate architecture from training recipe, and the stage hierarchy dominates: no stage can be removed without material loss, the least damaging omission still costing $7.61$ percentage points, while the parameter-matched controls add $3.17$, $1.63$, and $0.99$ points for augmentation, dropout, and Swish. No trimmed variant is a viable deployment alternative, which is why Section~\ref{sec:efficiency} relies on quantization rather than architectural reduction.\par
Section~\ref{sec:explainability} answers \textbf{RQ5}: the maps concentrate on the signing hand and suppress background even in cluttered classroom scenes, the deletion and insertion measures (deletion AUC $0.081$ versus $0.211$ for a random control) confirm that the highlighted region is the evidence actually used, and the randomization check confirms dependence on the learned parameters. For \textbf{RQ4}, the narrow 95.33 to 97.45\% band across ten models, including 20M-parameter backbones that saturate other BdSL benchmarks above 99\% \cite{tasnim2026vision,podder2022bangla}, makes RSBdSL38 a more challenging and ecologically valid benchmark than volunteer-collected datasets; the signer-independent drop of $11.19$ points to $85.18\%$ (Section~\ref{sec:signerindep}) reinforces this, a gap most stratified-split results leave unmeasured.\par
Several limitations remain. RSBdSL38 covers static alphabet signs only; dynamic word- and sentence-level signing needs temporal modeling outside this architecture's scope. On-device latency and memory are measured on a commodity smartphone SoC, but energy consumption and sustained-load thermal behaviour are not. The primary comparison uses stratified splits for consistency with the baselines and prior literature, and although the signer-independent evaluation quantifies the roughly 11-point optimism this introduces, the baselines are not re-evaluated under that protocol. The component study covers the training recipe and the activation function but not the individual attention modules, whose isolated contributions remain unquantified. Finally, the explainability analysis is limited by the $14 \times 14$ attribution resolution and is not complemented by a perturbation-based method such as SHAP \cite{lundberg2017unified}.
\section{Conclusion}
\label{sec:conclusion}
We introduced RSBdSL38, an expert-validated Bangla Sign Language dataset of 10,874 images collected from real signers across three regions of Bangladesh, together with a lightweight attention-based convolutional network of only 298{,}470 parameters (1.14~MB, 132.7~M multiply--accumulate operations). Trained from scratch it achieves 96.37\% accuracy on RSBdSL38 ($95.72\% \pm 0.54\%$ over five seeds), competitive with nine modern ImageNet-pretrained efficient architectures at 8.5 to 68$\times$ fewer parameters; it quantizes to a 0.48~MB INT8 model running at 3.98~ms per image on a commodity smartphone, generalizes across six public BdSL benchmarks (92.95 to 98.33\%) and a merged four-dataset corpus (97.04\%) without any architectural change, and transfers zero-shot to BdSL-38 at 76.25\%.\par
A signer-independent evaluation holding out 6 of the 36 signers entirely attains 85.18\%, quantifying the roughly 11-point optimism of the stratified splits standard in BdSL research. Two ablation studies separate architecture from training recipe, showing that no stage can be removed without material loss while augmentation, dropout, and the Swish activation each contribute a smaller but consistent part of the operating point. An explainability analysis using Grad-CAM \cite{selvaraju2017grad} and Grad-CAM++ \cite{chattopadhay2018grad}, validated by deletion and insertion faithfulness measures \cite{petsiuk2018rise} and a weight-randomization sanity check \cite{adebayo2018sanity}, confirms that the model localizes the signing hand rather than background context, including in cluttered classroom scenes.\par
Future work will extend the attribution analysis with perturbation-based methods such as SHAP \cite{lundberg2017unified}, isolate the channel and spatial attention modules through parameter-matched controls, extend RSBdSL38 to dynamic word-level signing and to a larger signer pool that supports signer-independent evaluation of the baselines as well, and characterize the energy and sustained-load thermal behaviour of the quantized model on target hardware. The dataset, trained models, and all experiment code are released publicly to support reproducible research on low-resource sign language recognition.
\section{Declaration of generative AI and AI-assisted technologies in the manuscript preparation process}
\label{sec:llm_usage}
Large language models (LLMs) were used during the preparation of this manuscript as assistive tools for language editing and code debugging. Every AI-assisted output was reviewed, verified, and corrected by the authors, who take full responsibility for the correctness, originality, and integrity of this work.\par
\paragraph{Originality} LLM assistance was editorial in nature. Drafted and refined passages were fact-checked and rewritten by the authors so that every statement matches the underlying research findings. All figures, tables, and quantitative results, including every technical description and reported numerical value, were produced entirely by the authors without AI involvement.\par
\paragraph{Transparency} LLMs were also used during code development, primarily to debug and locate implementation issues. The experimental pipeline nonetheless required substantial manual design, step-by-step correction, and repeated testing, as AI-generated suggestions alone were insufficient. Every reported result was obtained by running author-verified code and validated against expected behaviour before inclusion. The core methodology, the model architecture, and the RSBdSL38 experimental protocol were conceived and validated by the authors without LLM involvement.\par
\paragraph{Responsibility} No sensitive, private, or proprietary information, including dataset images and participant data, was shared with any AI tool during writing or debugging, and all interactions respected ethical considerations regarding data ownership and intellectual property. LLM use was confined to general-purpose writing assistance and code debugging, and did not influence the scientific contributions or claims of this paper.

\appendix
\setcounter{table}{0}
\setcounter{figure}{0}
\section{Summary of Notations}
\label{app:summaryofnotations}
Table ~\ref{tab:notation} provides a detailed summary of notations.
\begin{table*}[!t]
  \centering
  \caption{Summary of notation used throughout the paper.}
  \label{tab:notation}
  \setlength{\tabcolsep}{5pt}
  \renewcommand{\arraystretch}{1.25}
  \scriptsize
  \begin{tabularx}{\textwidth}{|c|X|c|X|}
    \hline
    \textbf{Symbol} & \textbf{Description} & \textbf{Symbol} & \textbf{Description} \\
    \hline
    $\mathbf{I}$ & Input RGB image, $\mathbf{I}\in\mathbb{R}^{224\times224\times3}$ & $\mathbf{X}_0$ & Stem output, $56\times56\times32$ \\
    \hline
    $\tilde{\mathbf{I}}$ & Augmented training image & $\mathbf{X}$ & Input of a residual block, $\mathbb{R}^{H\times W\times C}$ \\
    \hline
    $H,W,C$ & Height, width, channels of a feature map & $\mathbf{U},\mathbf{V}$ & Compressed and grouped bottleneck features \\
    \hline
    $H',W'$ & Size of the attribution grid ($14\times14$) & $\mathbf{Y}$ & Bottleneck output before attention \\
    \hline
    $K$ & Number of sign classes ($K=38$) & $\mathbf{Y}',\mathbf{Y}''$ & Channel- and spatially-refined features \\
    \hline
    $N$ & Number of training samples & $\mathbf{M}_c,\mathbf{M}_s$ & Channel and spatial attention maps \\
    \hline
    $B$ & Mini-batch size ($B=32$) & $\mathbf{Z}$ & Residual-block output \\
    \hline
    $i$ & Stage index, $i\in\{1,2,3,4\}$ & $\mathbf{Z}'_i$ & Output of stage $i$ \\
    \hline
    $C_i$ & Stage width, $C_i\in\{32,64,96,128\}$ & $\mathbf{H}$ & Projected input of the hand-feature block \\
    \hline
    $g_i$ & Convolution groups, $g_i\in\{2,4,8,16\}$ & $\mathbf{H}_3,\mathbf{H}_5$ & Outputs of the $3\times3$/$5\times5$ depthwise branches \\
    \hline
    $\rho_i$ & Spatial-dropout rate, $\rho_i\in\{0.05,\dots,0.20\}$ & $\mathbf{v}$ & Dual-pooled descriptor, $\mathbf{v}\in\mathbb{R}^{256}$ \\
    \hline
    $r$ & Channel-attention reduction ratio ($r=16$) & $\mathbf{h}_1,\mathbf{h}_2$ & Hidden activations, $\mathbb{R}^{128}$ and $\mathbb{R}^{64}$ \\
    \hline
    $p$ & Head dropout rate ($p=0.3$) & $\mathbf{W}_{\mathrm{stem}}$ & Stem kernel ($5\times5$, 32 filters) \\
    \hline
    $\lambda$ & $\ell_2$ regularization coefficient ($10^{-5}$) & $\mathbf{W}^{1\times1}_{c},\mathbf{W}^{1\times1}_{e}$ & Bottleneck compression and expansion kernels \\
    \hline
    $\ast$ & Standard (dense) convolution & $\mathbf{W}^{3\times3}_{g}$ & Grouped bottleneck kernel \\
    \hline
    $\ast_g$ & Grouped convolution with $g$ groups & $\mathbf{W}_{a0},\mathbf{W}_{a1}$ & Shared channel-attention MLP weights \\
    \hline
    $\ast_{s}$ & Convolution of stride $s$ & $f^{7\times7}$ & Spatial-attention convolution \\
    \hline
    $\mathrm{DW}^{k\times k}$ & Depthwise convolution with $k\times k$ kernel & $\mathbf{W}^{1\times1}_{p},\mathbf{W}^{1\times1}_{f}$ & Hand-feature projection and fusion kernels \\
    \hline
    $\mathrm{BN}(\cdot)$ & Batch normalization & $\mathbf{W}_{fc1},\mathbf{W}_{fc2}$ & Fully connected weight matrices \\
    \hline
    $\gamma,\beta,\epsilon$ & BN scale, shift, and numerical constant & $\mathbf{b}_1,\mathbf{b}_2$ & Fully connected bias vectors \\
    \hline
    $\mathrm{GAP},\mathrm{GMP}$ & Global average and global max pooling & $\mathbf{w}_k,b_k$ & Classifier weight vector and bias of class $k$ \\
    \hline
    $\mathrm{MaxPool}_{k,s}$ & Max pooling, window $k$, stride $s$ & $\boldsymbol{\theta}$ & All trainable parameters of the network \\
    \hline
    $\mathrm{SD}_{\rho}$ & Spatial (channel-wise) dropout at rate $\rho$ & $\mathbf{W}_l$ & Kernel of the $l$-th regularized layer \\
    \hline
    $\mathrm{Drop}_{p}$ & Element-wise dropout at rate $p$ & $\mathcal{L}$ & Regularized training objective \\
    \hline
    $\|$ & Channel-wise concatenation & $\eta_t,\eta_0,\eta_{\min}$ & Learning rate at step $t$, initial, and floor \\
    \hline
    $\otimes$ & Broadcast element-wise multiplication & $\mu,\mathbf{m}_t$ & Nesterov momentum and velocity buffer \\
    \hline
    $\circ$ & Function composition & $\tau_{\mathrm{geo}},\tau_{\mathrm{pho}}$ & Geometric and photometric augmentation operators \\
    \hline
    $\delta(\cdot)$ & Swish (SiLU) activation & $y_{nk}$ & One-hot target of sample $n$ for class $k$ \\
    \hline
    $\sigma(\cdot)$ & Logistic sigmoid & $\hat{y}_k,\hat{y}_{nk}$ & Predicted posterior probability \\
    \hline
    $\phi(\cdot)$ & ReLU, used inside the attention MLP & $TP,TN$ & True positives and true negatives \\
    \hline
    $\mathcal{P}(\cdot)$ & Residual shortcut projection & $FP,FN$ & False positives and false negatives \\
    \hline
    $\mathcal{S}_0,\mathcal{S}_i$ & Stem operator and stage-$i$ operator & $F_1$ & F1-score, harmonic mean of precision and recall \\
    \hline
    $\mathcal{B}^{(1)}_i,\mathcal{B}^{(2)}_i$ & First and second residual block of stage $i$ & $\Delta$ & Accuracy degradation w.r.t.\ baseline (pp) \\
    \hline
    $\mathcal{H}_i$ & Hand-feature block of stage $i$ & $A^k,y^c$ & $k$-th Stage-4 channel; pre-softmax score of class $c$ \\
    \hline
    $\pi_i$ & Stage transition (max pooling or identity) & $\alpha^c_k,L^c$ & Grad-CAM channel weight and localization map \\
    \hline
    $\mathcal{C}(\cdot),\mathcal{F}(\cdot)$ & Classifier head and complete network & $\mathrm{MACs}$ & Multiply--accumulate operations per forward pass \\
    \hline
  \end{tabularx}
\end{table*}

\section{Detailed Architecture}
\label{app:architecture}
\begin{figure*}[!t]
    \centering
    \includegraphics[width=0.8\textwidth]{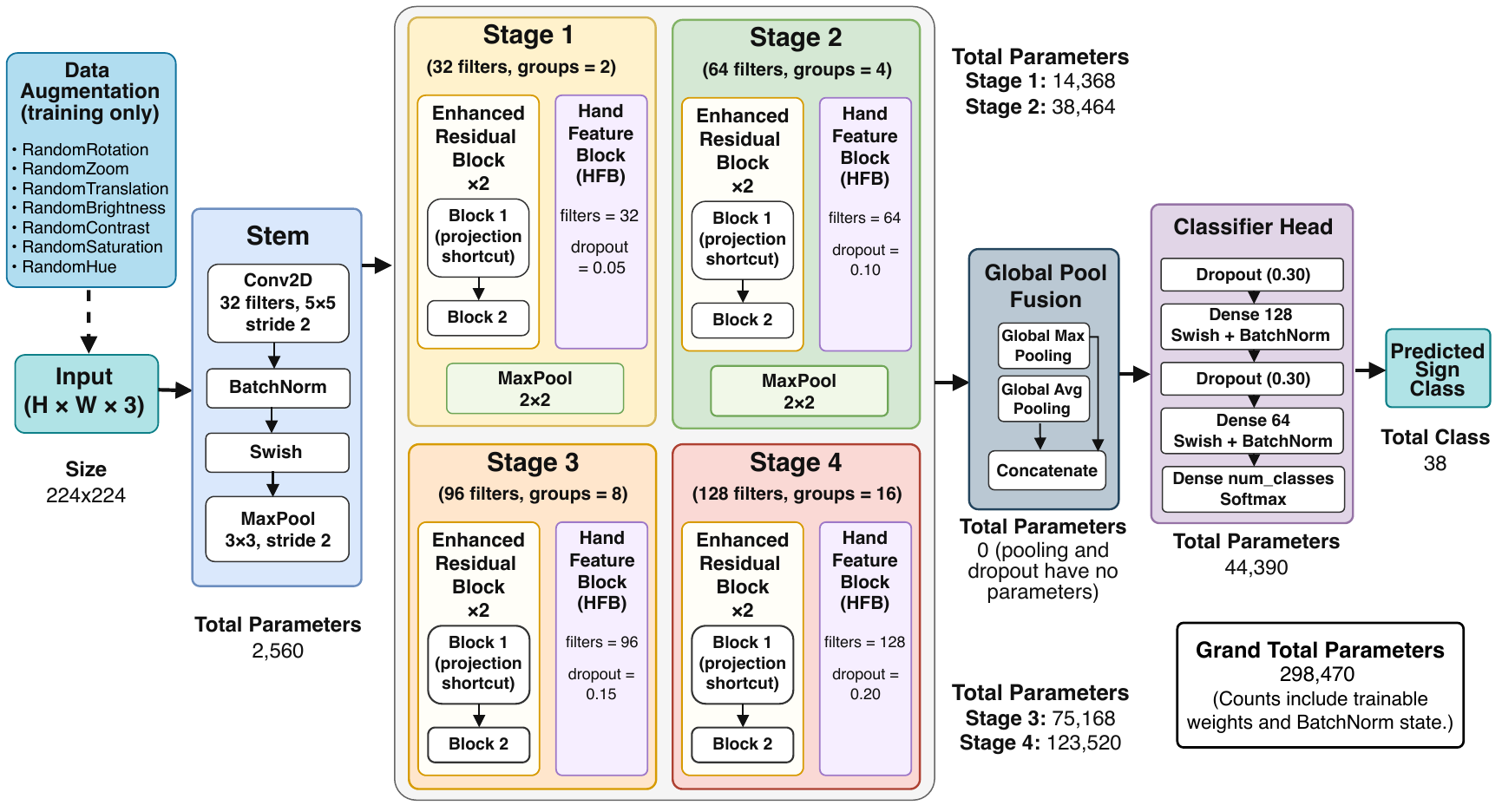}
    \caption{Layer-level architecture of the proposed model. The strided $5\times5$ stem is followed by four stages of increasing width (32, 64, 96, and 128 channels), each composed of two grouped bottleneck residual attention blocks and one multi-scale hand-feature block. Stages 1 and 2 terminate in $2\times2$ max pooling; stages 3 and 4 preserve the $14\times14$ resolution so that finger-level detail is not discarded. The dual-pooling head concatenates global max and global average descriptors into a 256-dimensional vector and maps it through two fully connected layers to the 38-way softmax.}
    \label{fig:model_arc_main}
\end{figure*}

\begin{figure*}[!t]
    \centering
    \begin{subfigure}[t]{0.9\textwidth}
        \centering
        \includegraphics[width=\linewidth]{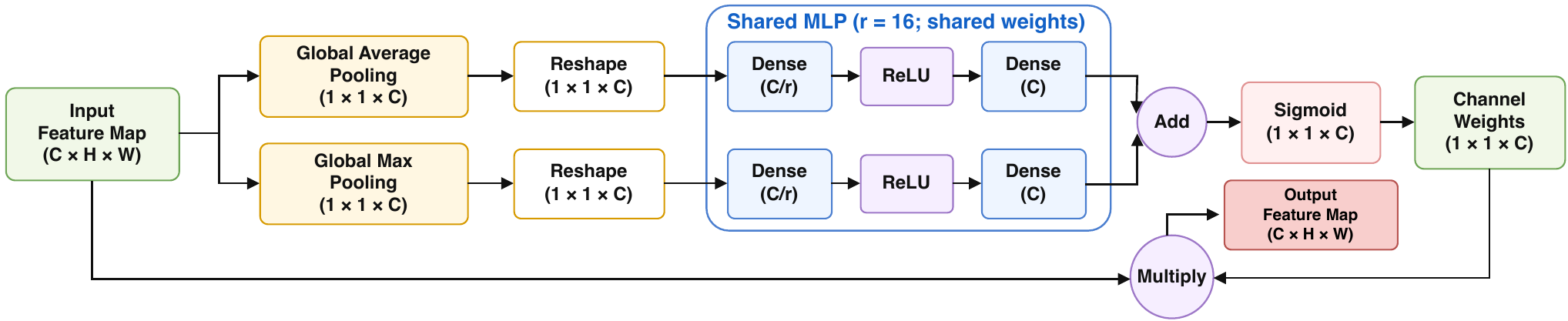}
        \caption{Channel-attention block. Global average- and max-pooled descriptors pass through a shared, bias-free two-layer MLP with reduction ratio $r=16$ to generate channel weights, which rescale the input feature map (Eq.~\eqref{eq:chatt}).}
        \label{fig:channel_attention}
    \end{subfigure}
    \vspace{2mm}
    \begin{subfigure}[t]{0.9\textwidth}
        \centering
        \includegraphics[width=\linewidth]{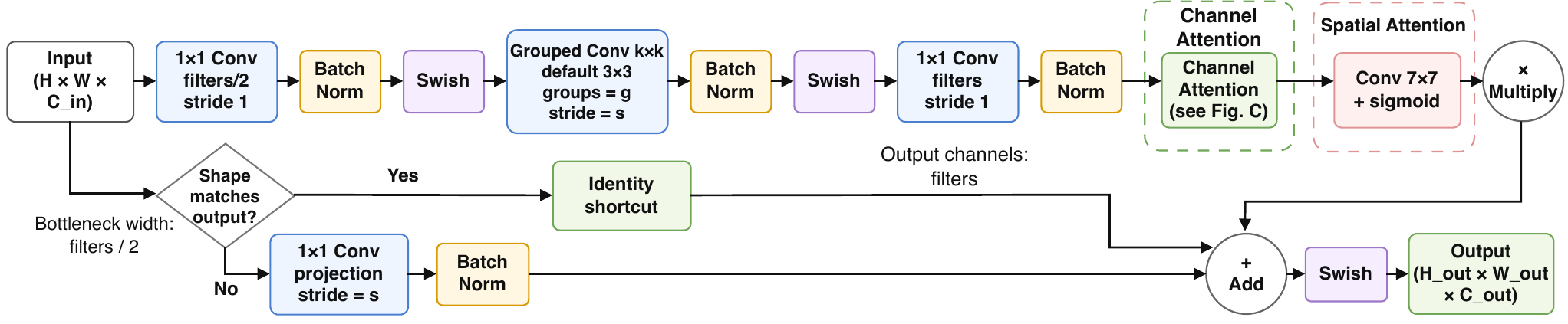}
        \caption{Grouped bottleneck residual attention block. A $1\times1$ compression, a grouped $3\times3$ convolution, and a $1\times1$ expansion are followed by channel and spatial attention, while an identity or projection shortcut supports residual learning (Eqs.~\eqref{eq:bneck1}--\eqref{eq:residual}).}
        \label{fig:enhanced_residual}
    \end{subfigure}
    \vspace{2mm}
    \begin{subfigure}[t]{0.9\textwidth}
        \centering
        \includegraphics[width=\linewidth]{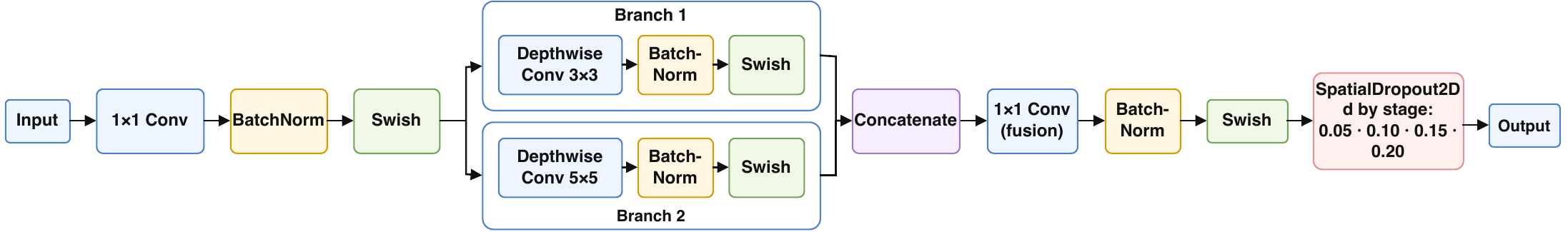}
        \caption{Hand-feature block. Parallel $3\times3$ and $5\times5$ depthwise convolutions capture finger-level and palm-level structure, and their outputs are concatenated, fused by a $1\times1$ convolution, and regularized by spatial dropout (Eqs.~\eqref{eq:hfb1}--\eqref{eq:hfb3}).}
        \label{fig:hand_feature}
    \end{subfigure}
    \caption{Internal structure of the three building blocks used in the proposed lightweight Bangla Sign Language recognition network: (a) the channel-attention block, (b) the grouped bottleneck residual attention block, and (c) the multi-scale hand-feature block.}
    \label{fig:proposed_building_blocks}
\end{figure*}

This appendix gives the layer-level view of the network summarized in Fig.~\ref{fig:model} and formalized in Section~\ref{sec:methodology}: Fig.~\ref{fig:model_arc_main} the complete computational graph, Fig.~\ref{fig:proposed_building_blocks} the internal structure of the three building blocks, and Table~\ref{tab:arch} the stage-wise configuration and parameter budget.\par

\begin{table}[!t]
  \centering
  \caption{Stage-wise configuration of the proposed model. Output size is the spatial resolution after the stage transition $\pi_i$; HFB denotes the multi-scale hand-feature block, $g_i$ the number of convolution groups, $\rho_i$ the spatial-dropout rate, and $p$ the head dropout rate.}
  \label{tab:arch}
  \setlength{\tabcolsep}{4pt}
  \scriptsize
  \begin{tabular}{@{}l c c c@{}}
    \toprule
    \textbf{Module} & \textbf{Output size} & \textbf{Width $C_i$} & \textbf{$g_i$ / $\rho_i$} \\
    \midrule
    Stem ($5\times5$, $s{=}2$, $+$ pool) & $56 \times 56$ & 32  & --- \\
    Stage 1 (2 blocks $+$ HFB)           & $28 \times 28$ & 32  & $g{=}2$, $\rho{=}0.05$ \\
    Stage 2 (2 blocks $+$ HFB)           & $14 \times 14$ & 64  & $g{=}4$, $\rho{=}0.10$ \\
    Stage 3 (2 blocks $+$ HFB)           & $14 \times 14$ & 96  & $g{=}8$, $\rho{=}0.15$ \\
    Stage 4 (2 blocks $+$ HFB)           & $14 \times 14$ & 128 & $g{=}16$, $\rho{=}0.20$ \\
    Dual-pool head (FC $128{\to}64{\to}38$) & $1 \times 1$ & --- & $p{=}0.3$ \\
    \midrule
    \multicolumn{4}{@{}l}{\textbf{Total: 298{,}470 parameters} ($\approx0.30$M; 292{,}262 trainable; 1.14~MB)} \\
    \bottomrule
  \end{tabular}
\end{table}
\section{Detailed Results on RSBdSL38}
\label{app:detailed}
\setcounter{table}{0}
\setcounter{figure}{0}
This appendix reports the per-seed stability statistics summarized in Section~\ref{sec:mainresults} and the complete per-class classification report of the single-seed reference run.\par
\begin{table}[!t]
\centering
\caption{Five-seed stability of the proposed model on RSBdSL38 (seeds 42--46), each retrained from scratch under the identical protocol. Epochs are the total run length before early stopping, and time is the wall-clock training time on a single GPU.}
\label{tab:multiseed}
\setlength{\tabcolsep}{4pt}
\footnotesize
\begin{tabular}{@{}l r r r r r r@{}}
\toprule
\textbf{Seed} & \textbf{Acc.} & \textbf{Prec.} & \textbf{Rec.} & \textbf{F1} & \textbf{Ep.} & \textbf{Time} \\
 & \textbf{(\%)} & \textbf{(\%)} & \textbf{(\%)} & \textbf{(\%)} & & \textbf{(min)} \\
\midrule
42 & 96.46 & 96.56 & 96.46 & 96.46 & 184 & 119.1 \\
43 & 95.92 & 96.00 & 95.92 & 95.91 & 184 & 118.9 \\
44 & 95.10 & 95.44 & 95.10 & 95.12 & 185 & 119.5 \\
45 & 95.83 & 96.03 & 95.83 & 95.83 & 215 & 145.0 \\
46 & 95.29 & 95.48 & 95.29 & 95.30 & 263 & 177.4 \\
\midrule
\textbf{Mean} & \textbf{95.72} & \textbf{95.90} & \textbf{95.72} & \textbf{95.72} & \textbf{206.2} & \textbf{136.0} \\
\textbf{Std}  & \textbf{0.54}  & \textbf{0.46}  & \textbf{0.54}  & \textbf{0.54}  & \textbf{34.4}  & \textbf{25.7} \\
Min  & 95.10 & 95.44 & 95.10 & 95.12 & 184 & 118.9 \\
Max  & 96.46 & 96.56 & 96.46 & 96.46 & 263 & 177.4 \\
\bottomrule
\end{tabular}
\end{table}
\begin{table*}[!t]
\centering
\caption{Per-class classification report of the proposed model on the RSBdSL38 test set (1{,}103 images, single-seed reference run). Support is the number of test images in each class. Four classes reach a perfect F1 of 1.00 and no class falls below 0.85 in either precision or recall.}
\label{tab:classreport}
\setlength{\tabcolsep}{5pt}
\footnotesize
\begin{tabular}{@{}c c c c c @{\hspace{16pt}} c c c c c@{}}
\toprule
\textbf{Class} & \textbf{Precision} & \textbf{Recall} & \textbf{F1-score} & \textbf{Support} &
\textbf{Class} & \textbf{Precision} & \textbf{Recall} & \textbf{F1-score} & \textbf{Support} \\
\midrule
0  & 0.97 & 1.00 & 0.98 & 30 & 19 & 0.93 & 0.93 & 0.93 & 30 \\
1  & 1.00 & 0.97 & 0.98 & 29 & 20 & 1.00 & 0.93 & 0.97 & 30 \\
2  & 1.00 & 0.93 & 0.96 & 27 & 21 & 1.00 & 1.00 & 1.00 & 30 \\
3  & 0.90 & 0.96 & 0.93 & 28 & 22 & 1.00 & 0.94 & 0.97 & 31 \\
4  & 0.93 & 0.96 & 0.95 & 27 & 23 & 0.94 & 1.00 & 0.97 & 29 \\
5  & 0.86 & 0.93 & 0.89 & 27 & 24 & 1.00 & 0.97 & 0.98 & 31 \\
6  & 0.93 & 0.93 & 0.93 & 29 & 25 & 0.97 & 1.00 & 0.98 & 30 \\
7  & 0.96 & 0.93 & 0.95 & 28 & 26 & 1.00 & 0.97 & 0.98 & 29 \\
8  & 0.93 & 1.00 & 0.97 & 28 & 27 & 1.00 & 0.93 & 0.96 & 28 \\
9  & 1.00 & 0.96 & 0.98 & 27 & 28 & 0.91 & 1.00 & 0.95 & 29 \\
10 & 0.94 & 0.94 & 0.94 & 31 & 29 & 1.00 & 0.96 & 0.98 & 28 \\
11 & 1.00 & 1.00 & 1.00 & 26 & 30 & 1.00 & 0.97 & 0.98 & 30 \\
12 & 0.89 & 0.86 & 0.87 & 28 & 31 & 1.00 & 0.97 & 0.98 & 29 \\
13 & 1.00 & 0.93 & 0.96 & 29 & 32 & 1.00 & 1.00 & 1.00 & 26 \\
14 & 0.97 & 0.90 & 0.93 & 31 & 33 & 0.97 & 0.97 & 0.97 & 31 \\
15 & 0.94 & 1.00 & 0.97 & 32 & 34 & 0.97 & 1.00 & 0.98 & 30 \\
16 & 0.96 & 0.96 & 0.96 & 28 & 35 & 0.97 & 1.00 & 0.98 & 32 \\
17 & 1.00 & 1.00 & 1.00 & 28 & 36 & 0.90 & 0.97 & 0.93 & 29 \\
18 & 0.94 & 1.00 & 0.97 & 29 & 37 & 1.00 & 0.97 & 0.98 & 29 \\
\midrule
\multicolumn{5}{@{}l}{\textbf{Accuracy}}     & \multicolumn{3}{c}{} & 0.96 & 1{,}103 \\
\multicolumn{5}{@{}l}{\textbf{Macro avg}}    & & 0.96 & 0.96 & 0.96 & 1{,}103 \\
\multicolumn{5}{@{}l}{\textbf{Weighted avg}} & & 0.97 & 0.96 & 0.96 & 1{,}103 \\
\bottomrule
\end{tabular}
\end{table*}
\section{Complete Ablation and Deployment Measurements}
\label{app:ablation}
\setcounter{table}{0}
\setcounter{figure}{0}
This appendix reports the complete per-variant metrics of the stage-wise depth ablation of Section~\ref{sec:ablationA}, the corresponding loss curves, and the full on-device measurement summarized in Section~\ref{sec:efficiency}.\par
\begin{table*}[!t]
  \centering
  \caption{Complete stage-wise depth ablation. D0 is the full-model baseline and every other row is an independently trained stage-removal configuration. Metrics are percentages and $\Delta$ is the accuracy degradation relative to D0, so larger values indicate a more damaging removal. Epochs are the total run length before early stopping.}
  \label{tab:stageablation}
  \setlength{\tabcolsep}{4pt}
  \footnotesize
  \begin{tabular}{@{}l l l r r r r r r r@{}}
    \toprule
    \textbf{Variant} & \textbf{Removed stages} & \textbf{Active stages} & \textbf{Params} & \textbf{Acc.} & \textbf{Prec.} & \textbf{Recall} & \textbf{F1} & \textbf{Epochs} & \textbf{$\Delta$ (pp)} \\
    \midrule
    \textbf{D0: Full model} & \textbf{None} & \textbf{S1+S2+S3+S4} & \textbf{298{,}470} & \textbf{96.55} & \textbf{96.65} & \textbf{96.55} & \textbf{96.55} & \textbf{192} & \textbf{0.00} \\
    \midrule
    D1  & S1          & S2+S3+S4 & 284{,}102 & 85.31 & 86.58 & 85.31 & 84.78 & 172 & $+11.24$ \\
    D2  & S2          & S1+S3+S4 & 255{,}398 & 88.94 & 90.75 & 88.94 & 88.64 & 221 &  $+7.61$ \\
    D3  & S3          & S1+S2+S4 & 217{,}158 & 85.22 & 88.03 & 85.22 & 84.67 & 172 & $+11.33$ \\
    D4  & S4          & S1+S2+S3 & 166{,}758 & 75.79 & 77.25 & 75.79 & 72.95 & 172 & $+20.76$ \\
    \midrule
    D5  & S1+S2       & S3+S4    & 241{,}030 & 79.87 & 83.40 & 79.87 & 78.02 & 156 & $+16.68$ \\
    D6  & S1+S3       & S2+S4    & 202{,}790 & 84.86 & 87.43 & 84.86 & 83.88 & 198 & $+11.69$ \\
    D7  & S1+S4       & S2+S3    & 152{,}390 & 82.96 & 87.34 & 82.96 & 81.33 & 198 & $+13.59$ \\
    D8  & S2+S3       & S1+S4    & 172{,}550 & 80.05 & 84.65 & 80.05 & 78.22 & 173 & $+16.50$ \\
    D9  & S2+S4       & S1+S3    & 123{,}686 & 83.05 & 86.87 & 83.05 & 81.56 & 171 & $+13.50$ \\
    D10 & S3+S4       & S1+S2    &  83{,}398 & 74.07 & 76.86 & 74.07 & 73.03 &  96 & $+22.48$ \\
    \midrule
    D11 & S1+S2+S3    & S4       & 158{,}182 & 80.15 & 82.16 & 80.15 & 78.81 & 114 & $+16.40$ \\
    D12 & S1+S3+S4    & S2       &  69{,}030 & 75.70 & 79.04 & 75.70 & 74.01 & 120 & $+20.85$ \\
    D13 & S2+S3+S4    & S1       &  36{,}742 & 31.10 & 29.17 & 31.10 & 26.06 &  43 & $+65.45$ \\
    D14 & S1+S2+S3+S4 & None     &  22{,}374 &  7.25 &  2.09 &  7.25 &  2.64 &  40 & $+89.30$ \\
    \bottomrule
  \end{tabular}
\end{table*}
\begin{figure*}[!t]
    \centering
    \includegraphics[width=0.8\textwidth]{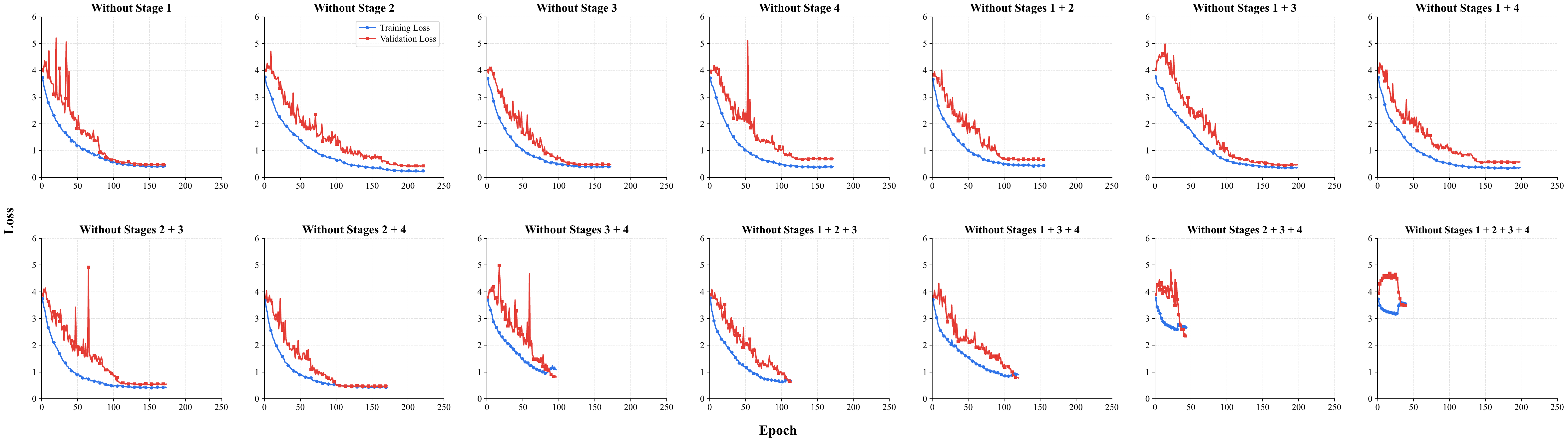}
    \caption{Training and validation loss for the 14 stage-removal configurations of Table~\ref{tab:stageablation}. All panels use identical epoch and loss scales, while each curve ends at its own completed training length; markers are drawn at regular intervals for legibility. The high residual losses of the most aggressively reduced variants agree with their large test-accuracy deficits.}
    \label{fig:ablationcurves}
\end{figure*}
\begin{table}[!t]
  \centering
  \caption{On-device inference of the full-integer INT8 model on a commodity Android smartphone (Snapdragon~7+~Gen~3, 12~GB RAM, XNNPACK delegate, 4 threads, 249 timed runs at $224\times224$).}
  \label{tab:ondevice}
  \setlength{\tabcolsep}{6pt}
  \footnotesize
  \begin{tabular}{@{}l r@{}}
    \toprule
    \textbf{Metric} & \textbf{Value} \\
    \midrule
    Model size (INT8 TFLite) & 0.48~MB \\
    Mean latency             & $3.98 \pm 0.09$~ms \\
    Median / p95 latency     & 3.96 / 4.16~ms \\
    Throughput               & $\approx$251~images/s \\
    Peak memory              & 15.5~MB \\
    Initialization time      & 13.8~ms \\
    Delegated operators      & 241 / 315 \\
    \bottomrule
  \end{tabular}
\end{table}

\section{Data and Code Availability}
\label{app:availability}
The public BdSL datasets used in this study are available from their original sources and are cited in the manuscript. The RSBdSL38 dataset introduced here is archived with a permanent DOI at \url{https://doi.org/10.17632/tgvmb2jsdb.1}. The trained weights, training histories, per-run metrics, and all code required to reproduce the reported results are available at \url{https://github.com/saadbaust/rsbdsl38}.
\bibliographystyle{elsarticle-harv}
\bibliography{references}
\end{document}